\pdfoutput=1

\documentclass[11pt]{article}

\usepackage{emnlp2021}

\usepackage{times}
\usepackage{latexsym}

\usepackage[T1]{fontenc}

\usepackage[utf8]{inputenc}

\usepackage{pgfplots}
\pgfplotsset{compat=1.14} 
\usepackage{graphicx}
\usepackage{xspace}
\usepackage{url}
\usepackage{pifont}
\usepackage{amsmath}
\usepackage{amssymb}
\usepackage{textcomp}
\usepackage{multirow}
\usepackage{graphicx}
\usepackage{microtype}
\usepackage{enumitem}
\usepackage{array}
\usepackage{lipsum} 
\usepackage{rotating}
\usepackage{xcolor,colortbl}
\usepackage{graphicx}
\usepackage{microtype}

\newcommand{\xlmroberta}{\textsc{xlm-roberta}\xspace}
\newcommand{\roberta}{\textsc{roberta}\xspace}
\newcommand{\bert}{\textsc{bert}\xspace}

\newcommand{\eu}{\textsc{eu}\xspace}
\newcommand{\nlp}{\textsc{nlp}\xspace}
\newcommand\eurlex{\textsc{eurlex57k}\xspace}
\newcommand{\multieurlex}{\textsc{multi-eurlex}\xspace}
\newcommand{\nativebert}{\textsc{native-bert}\xspace}
\newcommand{\transformer}{\textsc{transformer}\xspace}

\newcommand{\bitfit}{\textsc{bitfit}\xspace}
\newcommand{\lnfit}{\textsc{lnfit}\xspace}
\newcommand{\glue}{\textsc{glue}\xspace}
\newcommand{\eurovoc}{\textsc{eurovoc}\xspace}
\newcommand{\cls}{\texttt{\small [cls]}\xspace}
\newcommand{\s}{\texttt{\small </s>}\xspace}
\newcommand{\mt}{\textsc{mt5}\xspace}

\newcommand{\rnn}{\textsc{rnn}\xspace}

\newcommand\ffnn{\textsc{ffnn}\xspace}
\newcommand\mrp{m\textsc{rp}\xspace}
\definecolor{c15}{rgb}{0.05,0.04,0.03}
\definecolor{c22}{rgb}{0.10,0.073,0.09}
\definecolor{c24}{rgb}{0.1118,0.0747,0.0911}
\definecolor{c25}{rgb}{0.1281,0.0867,0.1140}
\definecolor{c26}{rgb}{0.1440,0.0998,0.1367}
\definecolor{c27}{rgb}{0.1597,0.1122,0.1595}
\definecolor{c28}{rgb}{0.1754,0.1240,0.1828}
\definecolor{c29}{rgb}{0.1945,0.1382,0.2126}
\definecolor{c30}{rgb}{0.2089,0.1495,0.2371}
\definecolor{c31}{rgb}{0.2224,0.1609,0.2622}
\definecolor{c32}{rgb}{0.2350,0.1724,0.2878}
\definecolor{c33}{rgb}{0.2466,0.1842,0.3138}
\definecolor{c34}{rgb}{0.2597,0.1992,0.3468}
\definecolor{c35}{rgb}{0.2691,0.2115,0.3736}
\definecolor{c36}{rgb}{0.2775,0.2241,0.4007}
\definecolor{c37}{rgb}{0.2850,0.2370,0.4281}
\definecolor{c38}{rgb}{0.2914,0.2501,0.4558}
\definecolor{c39}{rgb}{0.2979,0.2668,0.4910}
\definecolor{c40}{rgb}{0.3017,0.2806,0.5192}
\definecolor{c41}{rgb}{0.3038,0.2950,0.5467}
\definecolor{c42}{rgb}{0.3038,0.3104,0.5728}
\definecolor{c43}{rgb}{0.3015,0.3269,0.5964}
\definecolor{c44}{rgb}{0.2958,0.3489,0.6210}
\definecolor{c45}{rgb}{0.2897,0.3675,0.6362}
\definecolor{c46}{rgb}{0.2831,0.3864,0.6477}
\definecolor{c47}{rgb}{0.2769,0.4054,0.6564}
\definecolor{c48}{rgb}{0.2715,0.4242,0.6630}
\definecolor{c49}{rgb}{0.2660,0.4473,0.6694}
\definecolor{c50}{rgb}{0.2628,0.4655,0.6737}
\definecolor{c51}{rgb}{0.2606,0.4834,0.6775}
\definecolor{c52}{rgb}{0.2591,0.5012,0.6812}
\definecolor{c53}{rgb}{0.2578,0.5188,0.6851}
\definecolor{c54}{rgb}{0.2563,0.5407,0.6901}
\definecolor{c55}{rgb}{0.2552,0.5582,0.6942}
\definecolor{c56}{rgb}{0.2543,0.5757,0.6984}
\definecolor{c57}{rgb}{0.2537,0.5932,0.7025}
\definecolor{c58}{rgb}{0.2534,0.6107,0.7064}
\definecolor{c59}{rgb}{0.2538,0.6326,0.7112}
\definecolor{c60}{rgb}{0.2550,0.6501,0.7147}
\definecolor{c61}{rgb}{0.2572,0.6677,0.7179}
\definecolor{c62}{rgb}{0.2606,0.6852,0.7207}
\definecolor{c63}{rgb}{0.2657,0.7028,0.7232}
\definecolor{c64}{rgb}{0.2745,0.7246,0.7257}
\definecolor{c65}{rgb}{0.2839,0.7421,0.7272}
\definecolor{c66}{rgb}{0.2954,0.7595,0.7283}
\definecolor{c67}{rgb}{0.3087,0.7769,0.7289}
\definecolor{c68}{rgb}{0.3239,0.7942,0.7291}
\definecolor{c69}{rgb}{0.3472,0.8156,0.7287}
\definecolor{c70}{rgb}{0.3706,0.8323,0.7278}
\definecolor{c71}{rgb}{0.3996,0.8482,0.7268}
\definecolor{c72}{rgb}{0.4356,0.8630,0.7264}
\definecolor{c73}{rgb}{0.4785,0.8763,0.7280}
\definecolor{c74}{rgb}{0.5376,0.8909,0.7355}
\definecolor{c75}{rgb}{0.5844,0.9017,0.7456}
\definecolor{c76}{rgb}{0.6288,0.9125,0.7580}
\definecolor{c77}{rgb}{0.6711,0.9233,0.7731}
\definecolor{c78}{rgb}{0.7112,0.9341,0.7913}
\definecolor{c79}{rgb}{0.7582,0.9480,0.8181}
\definecolor{c80}{rgb}{0.7933,0.9595,0.8424}
\definecolor{c81}{rgb}{0.8264,0.9716,0.8685}
\definecolor{c82}{rgb}{0.8588,0.9839,0.8948}
\definecolor{c83}{rgb}{0.8906,0.9966,0.9210}

\title{MultiEURLEX -- A multi-lingual and multi-label legal document classification dataset for zero-shot cross-lingual transfer}

\author{Ilias Chalkidis$^{\;\dagger\;\diamond}$\qquad Manos Fergadiotis$^{\;\ddagger}$ \qquad \textbf{Ion Androutsopoulos$^{\;\ddagger}$} \\
$^{\dagger\;}$ Department of Computer Science, University of Copenhagen, Denmark \\ $^{\ddagger\;}$Department of Informatics, Athens University of Economics and Business, Greece \\
$^{\diamond\;}$Cognitiv+ Ltd., London, United Kingdom \\
\texttt{ilias.chalkidis@di.ku.dk} \qquad  \texttt{[fergadiotis,ion]@aueb.gr}}

\date{}

\begin{document}
\maketitle
\begin{abstract}
We introduce \multieurlex, a new multilingual dataset for topic classification of legal documents. The dataset comprises 65k European Union (\eu) laws, officially translated in 23 languages, annotated with multiple labels from the \eurovoc taxonomy. We highlight the effect of temporal concept drift and the importance of chronological, instead of random splits. We use the dataset as a testbed for zero-shot cross-lingual transfer,  where we exploit annotated training documents in one language (source) to classify documents in another language (target).
We find that fine-tuning a multilingually pretrained model (\xlmroberta, \mt) in a single source language leads to catastrophic forgetting of multilingual knowledge and, consequently, poor zero-shot transfer to other languages. Adaptation strategies, 
namely partial fine-tuning, adapters, \bitfit, \lnfit, originally proposed to accelerate fine-tuning for new end-tasks, help retain multilingual knowledge from pretraining, substantially improving zero-shot cross-lingual transfer, but their impact also depends on the pretrained model used and the size of the label set.
\end{abstract}

\section{Introduction}

Multilingual learning is an active field of research in \nlp. Starting from neural machine translation \cite{stahlberg2020}, multilingual neural models are increasingly being considered across \nlp tasks and multilingual benchmark datasets for cross-lingual language understanding are becoming available \cite{hu2020, ruder2021}, complementing previous monolingual benchmarks \cite{wang2018}. 
The initial paradigm of multilingual word embeddings \cite{ruder2017} was rapidly expanded to pretrained multilingual models \cite{conneau2018}, including work on zero-shot cross-lingual transfer \cite{Artetxe2019}. Multilingual models based on \transformer{s} \cite{vaswani2017}, jointly pretrained on large corpora across multiple languages, have significantly advanced the state-of-the-art in cross-lingual tasks \cite{conneau2020, xue2021}. 

\begin{figure}[t]
    \centering
    \includegraphics[width=\columnwidth]{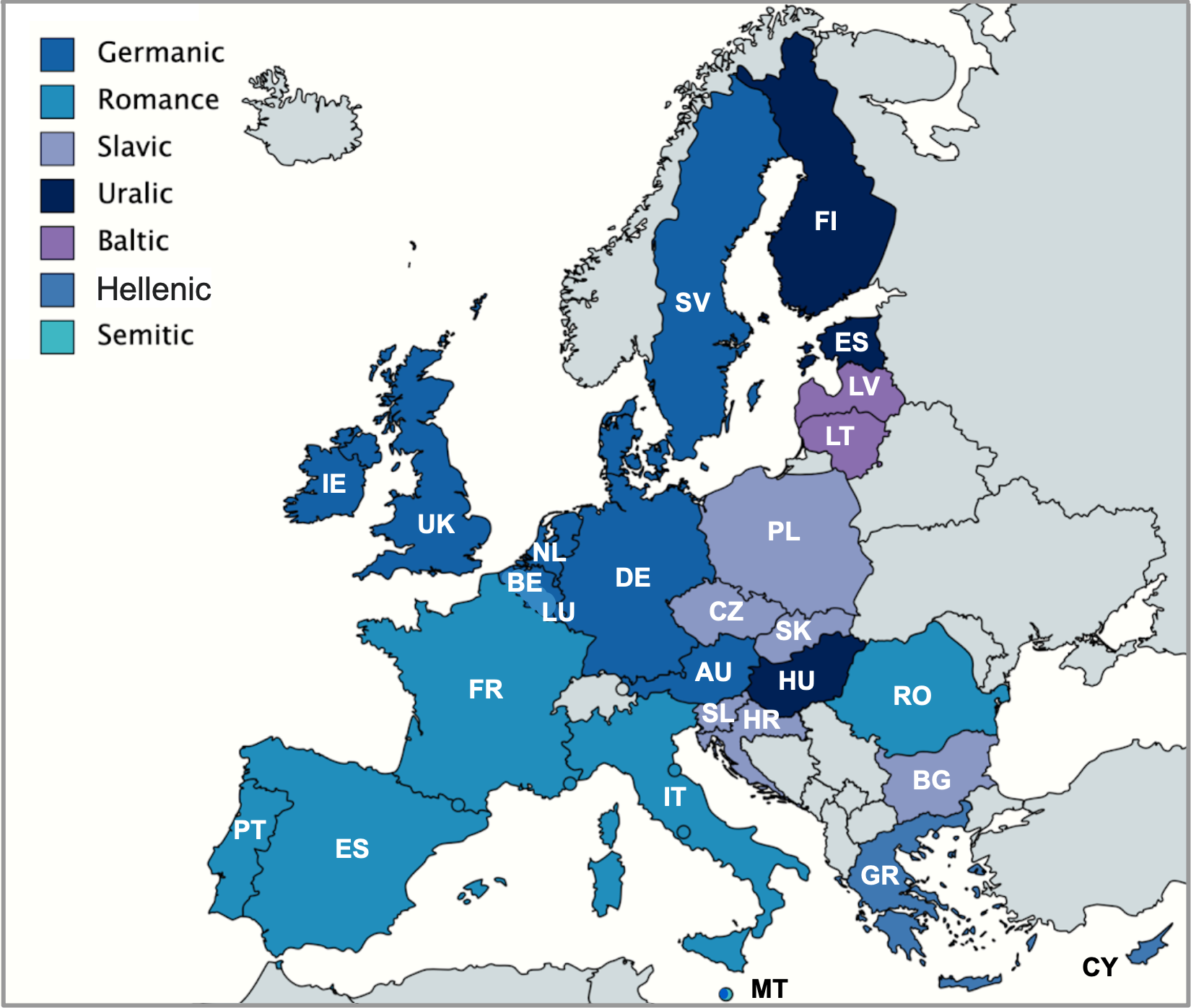}
    \vspace*{-6mm}
    \caption{\multieurlex covers 23 official \eu languages (Table~\ref{tab:stats}) from 7 families (illustrated per \eu country in the map). The \textsc{uk} was an \eu member until 2020. The map should not be taken to imply that no other languages are spoken in \eu countries.
    }
    \vspace*{-5mm}
    \label{fig:eu_languages}
\end{figure}

In another interesting direction, legal \nlp \citep{aletras2019,zhong2020} is an emerging field targeting  tasks such as legal judgment prediction \cite{Aletras2016}, legal topic classification \cite{chalkidis2019}, legal question answering \cite{Kim2015}, contract understanding \cite{hendrycks2021cuad}, to name a few. Generic pretrained language models for legal text in particular have also been introduced \cite{chalkidis2020b}. But despite rapid growth, cross-lingual transfer has not yet been explored in legal \nlp. 

\begin{figure}[t!]
    \centering
    \includegraphics[width=0.95\columnwidth]{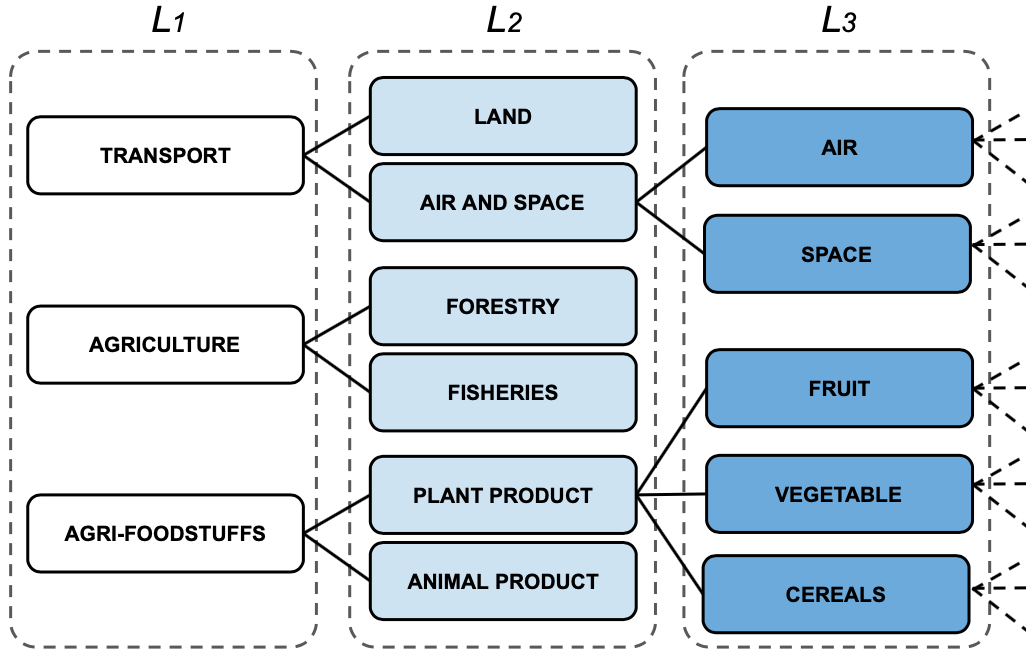}
    \vspace{-3mm}
    \caption{Examples from levels ($L_i$) 1 to 3 from the \eurovoc hierarchy. More general concepts become more specific as we move from higher to lower levels.}
    \label{fig:eurovoc}
    \vspace{-4mm}
\end{figure}

To facilitate research on cross-lingual transfer for text classification and legal topic classification in particular, we introduce a new multilingual dataset, \multieurlex, which includes 65k European Union (\eu) laws, officially translated  in the 23 \eu official languages (Fig.~\ref{fig:eu_languages}). Each document is annotated with multiple labels from \eurovoc, where concepts are organized hierarchically (Fig.~\ref{fig:eurovoc}).\footnote{\url{http://eurovoc.europa.eu/}} We use the dataset as a testbed for zero-shot cross-lingual transfer in cases where we wish to exploit labeled training documents in one language (source) to classify documents in another language (target).
This would allow, e.g., classifiers trained in resource-rich languages to be reused in languages with fewer or no training instances. 

We experiment with monolingual and multilingual \transformer{-based} models, i.e., monolingual \bert models \cite{devlin2019}, \xlmroberta \cite{conneau2020}, and \mt \cite{xue2021}.
We find that fine-tuning a multilingual model in a single source language leads to catastrophic forgetting of multilingual knowledge and, consequently, poor zero-shot transfer to target languages. 
We show that adaptation strategies, namely, not fine-tuning some layers, adapters \cite{houlsby2019}, \bitfit \cite{zaken2021}, and \lnfit inspired by \citet{frankle2021}, originally proposed to accelerate fine-tuning for new end-tasks, help retain multilingual knowledge from pretraining, substantially improving zero-shot cross-lingual transfer, but their impact also depends on the particular pretrained model used and the size of the label set.
We also compare chronological vs.\ random splits, highlighting the impact of temporal concept drift in legal topic classification, which causes random splits to over-estimate performance \cite{sogaard2021}.
Our main contributions are: 

\begin{itemize}[leftmargin=12pt]
\vspace*{-2mm}
    \setlength\itemsep{-1mm}
    \item A parallel multilingual annotated dataset for legal topic classification with 65k \eu laws in 23 languages, which can be used as a testbed for cross-lingual multi-label classification.
    
    \item Extensive experiments with state-of-the-art monolingual and multilingual models in 23 languages, which establish strong baselines for research on cross-lingual (legal) text classification.
    
    \item Experiments with several adaptation strategies 
    showing that adaptation is beneficial in zero-shot cross-lingual transfer, apart from task transfer.
    
    \item Comparison of chronological vs.\ random 
    splits, showing the temporal concept drift in legal topic classification and problems with random splits.\vspace{-4mm}

\end{itemize}

\begin{table*}[t]
    \centering
    \resizebox{\textwidth}{!}{
    {\small
    \begin{tabular}{l|c|l|c|c|c|c|c|r}
        \hline
        \multirow{2}{*}{\bf Language} & \bf ISO & \multirow{2}{*}{\bf Member Countries where official} & \multicolumn{2}{c|}{\bf EU Speakers (\%)} &\multicolumn{3}{c|}{\bf Number of Documents} & \bf \multirow{2}{*}{\shortstack{Words per\\ document}} \\
        & \bf code &  & Native & Total & Train & Dev. & Test &  \\
        \hline
        \multirow{2}{*}{English}     & \multirow{2}{*}{\textbf{en}}   & \footnotesize United Kingdom (1973--2020),   & \multirow{2}{*}{13\%} & \multirow{2}{*}{51\%} & \multirow{2}{*}{55,000} & \multirow{2}{*}{5,000} & \multirow{2}{*}{5,000} &  \multirow{2}{*}{1200 / 460} \\
        & & \footnotesize Ireland (1973), Malta (2004) & & & & & \\
        \multirow{2}{*}{German}      & \multirow{2}{*}{\textbf{de}}   & \footnotesize Germany (1958), Belgium (1958),  & \multirow{2}{*}{16\%} & \multirow{2}{*}{32\%} & \multirow{2}{*}{55,000} & \multirow{2}{*}{5,000} & \multirow{2}{*}{5,000} &  \multirow{2}{*}{1085 / 410} \\
         & & \footnotesize Luxembourg (1958) & & & & & \\
        \multirow{2}{*}{French}      & \multirow{2}{*}{\textbf{fr}}   & \footnotesize France (1958), Belgium(1958),  & \multirow{2}{*}{12\%} & \multirow{2}{*}{26\%} & \multirow{2}{*}{55,000} & \multirow{2}{*}{5,000} & \multirow{2}{*}{5,000} &  \multirow{2}{*}{1280 / 480} \\
         & & \footnotesize Luxembourg (1958) & & & & & \\
        Italian     & \textbf{it}   & \footnotesize Italy (1958)   & 13\% & 16\% & 55,000 & 5,000 & 5,000 & 1210 / 460 \\
        Spanish     & \textbf{es}   & \footnotesize Spain (1986)  & 8\% & 15\% & 52,785 & 5,000 & 5,000 & 1380 / 530 \\
        Polish      & \textbf{pl}   & \footnotesize Poland (2004)  & 8\% & 9\% & 23,197 & 5,000 & 5,000 & 1200 / 420 \\
        Romanian    & \textbf{ro}   & \footnotesize Romania (2007)  & 5\% & 5\% & 15,921 & 5,000 & 5,000 & 1500 / 500 \\
        Dutch       & \textbf{nl}   & \footnotesize Netherlands (1958), Belgium (1958)  & 4\% & 5\% & 55,000 & 5,000 & 5,000 &  1230 / 470\\
        Greek       & \textbf{el}   & \footnotesize Greece (1981), Cyprus (2008) & 3\% & 4\% & 55,000 & 5,000 & 5,000 & 1230 / 470 \\
        Hungarian   & \textbf{hu}   & \footnotesize Hungary (2004)  & 3\% & 3\% & 22,664 & 5,000 & 5,000 & 1120 / 370 \\
        Portuguese  & \textbf{pt}   & \footnotesize Portugal (1986)  & 2\% & 3\% & 23,188 & 5,000 & 5,000 & 1290 / 500 \\
        Czech       & \textbf{cs}   & \footnotesize Czech Republic (2004)  & 2\% & 3\% & 23,187 & 5,000 & 5,000 & 1170 / 410\\
        Swedish     & \textbf{sv}   & \footnotesize Sweden (1995)  & 2\% & 3\% & 42,490 & 5,000 & 5,000 & 1130 / 470\\
        Bulgarian   & \textbf{bg}   & \footnotesize Bulgaria (2007)  & 2\% & 2\% & 15,986 & 5,000 & 5,000 & 1480 / 510 \\
        Danish      & \textbf{da}   & \footnotesize Denmark (1973)  & 1\% & 1\% & 55,000 & 5,000 & 5,000 & 1080 / 410 \\
        Finnish     & \textbf{fi}   & \footnotesize Finland (1995)  & 1\% & 1\% & 42,497 & 5,000 & 5,000 & 890 / 320\\
        Slovak      & \textbf{sk}   & \footnotesize Slovakia (2004)  & 1\% & 1\% & 15,986 & 5,000 & 5,000 & 1180 / 410 \\
        Lithuanian  & \textbf{lt}   & \footnotesize Lithuania (2004)   & 1\% & 1\% & 23,188 & 5,000 & 5,000 & 1070 / 370 \\
        Croatian    & \textbf{hr}   & \footnotesize Croatia (2013) & 1\% & 1\% & 7,944 & 2,500 & 5,000 & 1490 / 500  \\
        Slovene     & \textbf{sl}   & \footnotesize Slovenia (2004) & <1\% & <1\% & 23,184 & 5,000 & 5,000 & 1170 / 400 \\
        Estonian    & \textbf{et}   & \footnotesize Estonia (2004) & <1\% & <1\% & 23,126 & 5,000 & 5,000 & 950 / 330 \\
        Latvian     & \textbf{lv}   & \footnotesize Latvia (2004) & <1\% & <1\% & 23,188 & 5,000 & 5,000 & 1080 / 380 \\
        Maltese     & \textbf{mt}   & \footnotesize Malta (2004) & <1\% & <1\% & 17,521 & 5,000 & 5,000 & 1250 / 430 \\
        \hline
        
    \end{tabular}
    }}
    \vspace*{-3mm}
    \caption{\multieurlex statistics per language: \textsc{iso} code; \eu countries using the language officially (year the country joined the \eu in brackets); percentage of \eu population speaking the language natively or in total (as native or non-native speakers);\footnotemark[3] 
    documents in training, development, test splits; words per document (mean/median).
    }
    \label{tab:stats}
    \vspace{-3mm}
\end{table*}

\section{Related Work}
\vspace*{-1mm}
Legal topic classification has been studied for \eu legislation \cite{Mencia2007,chalkidis2019} in a monolingual setting (English). While there are several legal \nlp studies with non-English datasets \cite{Kim2015,Waltl2017, Truong2017, Angelidis2018, dearaujo2020}, cross-lingual transfer has not been studied in the legal domain. 

Cross-lingual transfer is a very active area of wider \nlp research, currently dominated by large multilingually pretrained models 
\cite{conneau2018, eisenschlos2019, liu2020, xue2021}. Recent work explores adapter modules  
\cite{houlsby2019} to transfer monolingually pretrained \cite{artetxe2020} or multilingually pretrained \cite{pfeiffer2020} models to new (target) languages.
We examine more adaptation strategies, apart from adapter modules, in truly zero-shot 
cross-lingual transfer. 
Unlike \citet{pfeiffer2020}, we do not train language-specific adapters per target language; we use adapters to fine-tune a \emph{single} multilingual model on the source language, which is then used in all target languages.

In the broader field of multilingual legal studies, \citet{gonalves2010} examined legal topic classification with a dataset comprising 2.7k \eu laws in 4 languages (English, German, Spanish, Portuguese). They experimented with monolingual \textsc{svm} classifiers and their combination as a multilingual ensemble. More recently, \citet{galassi2020} transferred sentence-level gold labels from annotated  English to non-annotated German sentences, for the task of identifying unfair clauses in Terms of Service (2.7k sentences) and Privacy Policy documents (1.8k). They experimented with similarity-based methods aligning the English sentences to machine-translations of the German sentences.
We experiment with state-of-the-art multilingual \transformer{-based} models considering many more languages (23) and a much larger dataset (65k \eu laws).
Although \multieurlex is largely parallel, we use it as a testbed for zero-shot cross-lingual transfer, \emph{without} requiring parallel training data or machine translation systems.

\section{The \multieurlex Dataset \footnote{The dataset is available at \url{https://huggingface.co/datasets/multi_eurlex}. Following \citet{gebru2018}, we provide an extended 
Dataset Card in Appendix~\ref{sec:data_card}.}} \label{sec:dataset}

\noindent{\textbf{Documents:}}
\multieurlex comprises 65k \eu laws in 23 official \eu languages (Table~\ref{tab:stats}). Each \eu law has been annotated with \eurovoc concepts (labels) by the Publications Office of \eu. Each \eurovoc label \textsc{id} is associated with a \emph{label descriptor}, e.g., $\langle$60, `agri-foodstuffs'$\rangle$,  $\langle$6006, `plant product'$\rangle$, $\langle$1115, `fruit'$\rangle$. The descriptors are also available in the 23 languages. \citet{chalkidis2019} published a \emph{monolingual} (English) version of this dataset, called \eurlex, comprising 57k \eu laws with the originally assigned gold labels.\vspace{2mm}

\noindent{\textbf{Languages:}} \multieurlex covers 23 languages from 7 families (Fig.~\ref{fig:eu_languages}). \eu laws are published in all official \eu languages, except for Irish for resource-related reasons.\footnote{\url{https://europa.eu/european-union/about-eu/eu-languages_en}} This wide coverage makes the dataset a valuable testbed for cross-lingual transfer. All languages use the Latin script, except for Bulgarian (Cyrillic script) and Greek.\vspace{2mm}
\footnotetext[4]{Data from \citet{ec2012}. Following \textsc{brexit} (2020), \textsc{uk} citizens are no longer considered \eu citizens, thus native English speakers became approx.~1\% in the \eu, as of 2021. Table~\ref{tab:stats} includes \textsc{uk} citizens.} 
\addtocounter{footnote}{1}

\begin{table}[h]
    \centering
    \resizebox{\columnwidth}{!}{
    \begin{tabular}{c|c|rr|rr}
        \hline 
         \bf Label Set & \bf No. of Labels & \multicolumn{2}{c|}{\bf In training docs} & \multicolumn{2}{c}{\bf In all docs}   \\
         \hline
         Level 1 &  21 & 21 & (100\%) & 21 & (100\%) \\
         Level 2 &  127 & 127 & (100\%)  & 127 & (100\%) \\
         Level 3 &  567 & 500 & (88\%) & 511 & (90\%) \\
         All & 7,390 & 4,220 & (57\%) & 4,591 & (62\%)\\
         \hline
    \end{tabular}
    }
    \vspace{-3mm}
    \caption{\eurovoc concepts in the four label sets and how many are used in the training or entire dataset.}
    \label{tab:eurovoc}
    \vspace{-4mm}
\end{table}

\noindent\textbf{Multi-granular Labeling:} 
\eurovoc has eight levels of concepts (Fig.~\ref{fig:eurovoc} illustrates three). Each document is assigned one or more concepts (labels). If a document is assigned a concept, the ancestors and descendants of that concept are typically not assigned to the same document. The documents were originally annotated with concepts from levels 3 to 8. We created three alternative sets of labels per document, by replacing each assigned concept by its ancestor from level 1, 2, or 3, respectively. Thus, we provide four sets of gold labels per document, one for each of the first three levels of the hierarchy, plus the original sparse label assignment.\footnote{Levels 4 to 8 cannot be used independently, as many documents have gold concepts from the third level; thus many documents will be mislabeled, if we discard level 3.} Table~\ref{tab:eurovoc} presents the distribution of labels across label sets.\vspace{2mm}

\noindent{\textbf{Supported Tasks:}} 
Similarly to \eurlex \cite{chalkidis2019}, \multieurlex can be used for legal topic classification, a multi-label classification task where legal documents need to be assigned concepts (in our case, from \eurovoc) reflecting their topics. Unlike \eurlex, however, \multieurlex supports labels from three different granularities (\eurovoc levels). More importantly, apart from monolingual (\emph{one-to-one}) experiments, it can be used to study cross-lingual transfer scenarios, including \emph{one-to-many} (systems trained in one language and used in other languages with no training data), and \emph{many-to-one} or \emph{many-to-many} (systems jointly trained in multiple languages and used in one or more other languages).\vspace{2mm}

\noindent{\textbf{Data Split and Concept Drift:}} 
\multieurlex is \emph{chronologically} split in training (55k, 1958--2010), development (5k, 2010--2012), test (5k, 2012--2016) subsets, using the English documents. The test subset contains the same 5k documents in all 23 languages (Table~\ref{tab:stats}).\footnote{The development subset also contains the same 5k documents in 23 languages, except Croatian. Croatia is the most recent \eu member (2013); older laws are gradually translated.}
For the official languages of the seven oldest member countries, the same 55k training documents are available; for the other languages, only a subset of the 55k training documents is available (Table~\ref{tab:stats}).
Compared to \eurlex \cite{chalkidis2019}, \multieurlex is not only larger (8k more documents) and multilingual; it is also more challenging, as the chronological split leads to temporal real-world \emph{concept drift} across the training, development, test subsets, i.e., differences in label distribution and phrasing, representing a realistic \emph{temporal generalization} problem \cite{huang2019,lazaridou2021}. Recently, \citet{sogaard2021} showed this setup is more realistic, as it does not over-estimate real performance, contrary to random splits \cite{gorman-bedrick-2019-need}.

\begin{table}[h]
    \centering
    \resizebox{\columnwidth}{!}{
    \begin{tabular}{c|c|c|c|c}
         \hline
         \multirow{2}{*}{\bf Label Set} &  \multicolumn{2}{c|}{\bf Random }& \multicolumn{2}{c}{\bf Chronological} \\
         & \emph{train-dev} & \emph{train-test} & \emph{train-dev} & \emph{train-test} \\
         \hline
         Level 1 & 0.00 & 0.00 & \cellcolor{blue!4}0.03 & \cellcolor{blue!5} 0.04 \\
         Level 2 & 0.00 & 0.00 & \cellcolor{blue!12} 0.12 & \cellcolor{blue!16} 0.16 \\
         Level 3 & \cellcolor{blue!1} 0.01 & \cellcolor{blue!1} 0.01 & \cellcolor{blue!21} 0.21 & \cellcolor{blue!32} 0.32 \\
         All & \cellcolor{blue!20} 0.20 & \cellcolor{blue!20} 0.20 & \cellcolor{blue!60} 1.09 & \cellcolor{blue!75} \textcolor{white}{1.67} \\ 
         \hline
    \end{tabular}
    }
    \vspace*{-3mm}
    \caption{\textsc{kl}-divergence of label distributions between subsets, using a \emph{random} or \emph{chronological} split.}
    \label{tab:kl}
    \vspace{-3mm}
\end{table}

To verify that the chronological split of \multieurlex in training, development, test subsets leads to a \emph{temporal concept drift}, we compare the \textsc{kl}-divergence between the label distributions of the subsets using the chronological vs.\ a random split. Table~\ref{tab:kl} shows a random split leads to almost zero divergence for levels 1--3 and low divergence when using all labels. With the chronological split, the divergence increases as the number of labels increases, and is larger between the train and test subsets, which have a larger temporal distance compared to the train and development subsets. 

\begin{table}[h]
    \centering
    \vspace{-1mm}
    \resizebox{0.9\columnwidth}{!}{
    \begin{tabular}{l|c|c|c}
    \hline
    \bf Data Split & Training & Development & Test \\
    \hline
    Random &  \bf 99.2 &  \bf 74.7 &  \bf 74.0 \\
    Chronological & 96.7 &  58.7 &  48.4 \\
    \hline
    \end{tabular}
    }
     \vspace{-2mm}
    \caption{Results of \multieurlex for the original sparse annotation (7,390 labels) with \bert using a \emph{random} or \emph{chronological} split. Here the model is fine-tuned and tested on English data only (\emph{one-to-one}).}
    \vspace{-3mm}
    \label{fig:data_splits}
\end{table}

To further highlight the temporal concept drift, we fine-tune \bert \cite{devlin2019} on the English part of \multieurlex using all labels, following \citet{chalkidis2019}. Table~\ref{fig:data_splits} shows that although the performance on training data is very high with both splits, it deteriorates more rapidly on development data with the chronological split. Also, performance is stable when moving from development to test data with the random split, since both subsets contain randomly sampled unseen documents; but with the chronological split, performance continues to decline on test data. This confirms our hypothesis of a temporal concept drift and shows that the random split over-estimates real performance, contrary to the chronological split.

\section{Methods} \label{sec:methods}

\vspace*{-1mm}
\subsection{Pretrained Models}

\vspace*{-1mm}
\noindent\textbf{\nativebert{s}:} Many monolingual pretrained \transformer{-based} \cite{vaswani2017} models have been released, based on \bert \cite{devlin2019} or \roberta \cite{liu2019}.\footnote{Appendix~\ref{sec:appendix_c} lists the native pretrained \bert{s} we used.} Across classification experiments, $L$ is the cardinality of the label set, and $D_h$ the dimensionality of the hidden states. We feed the top-level hidden state of the \cls token ($\in\mathbb{R}^{D_h}$) to a dense layer  ($W_{\cls}\in\mathbb{R}^{D_h\times L}$) with $L$ outputs and sigmoids.\vspace{2mm}

\noindent\textbf{\xlmroberta:} \citet{conneau2020} introduced a multilingual \roberta for 100 languages. It is pretrained on Common Crawl with a vocabulary of 250k sub-words shared across languages. We use the same classification setup as in \nativebert{s}.\vspace{2mm}

\noindent\textbf{\mt:} 
\citet{xue2021} released a multilingual variant of \textsc{t5} \cite{raffel2020}, an encoder-decoder \transformer-based model, pretrained on text in 101 languages from Common Crawl. 
As in \textsc{t5}, \citet{xue2021} frame all \nlp tasks (incl. text classification) as text generation. This approach (text-to-text) is reasonable in single-label multi-class classification tasks like those of \glue \cite{wang2018}, where the output is expected to be the textual descriptor of a single class. But in our case, we have a \emph{multi-label} task with 5 labels per document on average and label sets containing hundreds or thousands of labels; hence a textual output would be unnecessarily complex. 
Also, requiring a \emph{sequence} of labels as output would be problematic, since the correct labels are not ordered. Hence, we use only the encoder of \mt. Similarly to \xlmroberta, we add a \cls special token, always at the beginning of the sequence, and use its top-level hidden state to represent the document.\footnote{In additional experiments, we also examined the original generative fine-tuning of \mt \cite{xue2021}, and another simplified encoder-decoder variant of \mt agnostic of label order. Both led to worse performance, while being substantially larger (40\% more parameters). See Appendix~\ref{sec:appendix_mt}.}

\subsection{Cross-lingual Adaptation Strategies}

We mainly study \emph{zero-shot cross-lingual transfer}, where we fine-tune (further train) a multilingual model (pretrained on a multilingual corpus) only on annotated documents of a \emph{source} language, and evaluate it (without any further training) on test (and development) documents in the other 22 languages (\emph{one-to-many}). To avoid \emph{catastrophically forgetting} the multilingual pretraining when fine-tuning only for the source language, we examine adaptation strategies, where the model is only partially fine-tuned. 
These were originally proposed to accelerate fine-tuning when moving to new end-tasks, but we employ them to retain multilingual knowledge. The four strategies are the following:\vspace{2mm}

\noindent\textbf{Frozen layers:} 
In this case, we follow \citet{rosenfeld2018} and do not update the parameters of the first $N$ or all ($N=$12) stacked \transformer blocks in fine-tuning; we also never update any input embeddings (of tokens, positions, segments). We experiment with $N=$ 3, 6, 9, 12.\vspace{2mm}

\noindent\textbf{Adapter modules:} 
In this case, we follow \citet{houlsby2019}, placing adapter modules after each feed-forward layer (\ffnn) inside each \transformer encoder block. Each block contains two \ffnn layers: one after the attention layer and one at the very end. 
An adapter module consists of a down-projection dense layer ($W_{down} \in \mathbb{R}^{D_h \times K}$, assuming row-vectors, where $K \ll D_h$) and a consecutive up-projection ($W_{up} \in \mathbb{R}^{K\times D_h}$), followed by a residual connection \citep{he2016}. The rest of the Transformer block is not updated, except for layer normalization components \cite{ba2016}.\vspace{2mm}

\noindent\textbf{BitFit:}
\bitfit \cite{zaken2021} keeps the whole network frozen during fine-tuning, except for bias terms. \citeauthor{zaken2021} showed that applying \bitfit on the English \bert (updating 0.09\% of parameters) is competitive with fully fine-tuning the entire model in the \glue benchmark \cite{wang2018}.\vspace{2mm}

\noindent\textbf{LNFit:}
Similarly, \citet{frankle2021} train only the parameters of \emph{batch normalization} \cite{ioffe2015} layers in image classifiers. We adopt a similar approach, dubbed \lnfit, where we fine-tune only the \emph{layer normalization} parameters of pre-trained \transformer{s} for text.
\vspace{1mm}

\noindent 
The randomly-initialized classification (dense) layer on top of the encoder is always fine-tuned.\vspace*{-1mm}

\section{Experimental Setup}
\label{sec:setup}

\noindent\textbf{Configuration of Models and Training Details:}  We implemented all methods in \textsc{tensorflow~2}, obtaining pretrained models from the Hugging Face library. We release our code and data for reproducibility.\footnote{Our code is available on Github (\url{https://github.com/nlpaueb/multi-eurlex}).} 
All models follow the \textsc{base} configuration with 12 stacked \transformer encoder blocks, each with $D_h=768$ and 12 attention heads. We use the Adam optimizer \citep{Kingma2015} across all experiments. We grid-search to tune the learning rate per method, considering classification performance on development data.\footnote{See Appendix~\ref{sec:appendix_e} for details on hyper-parameter tuning.}\vspace{2mm}

\begin{table*}[t]
    \centering
    \resizebox{\textwidth}{!}{
    \begin{tabular}{lc|cccc|ccccc|ccc|cc|c|c}
        \hline
           & \multicolumn{5}{c}{\textsc{Germanic}} & \multicolumn{5}{c}{\textsc{Romance}} & \multicolumn{3}{c}{\textsc{Slavic}} & \multicolumn{2}{c}{\textsc{Uralic}} & \multicolumn{2}{c}{} \\
          & \bf en & \bf da & \bf de & \bf nl & \bf sv & \bf ro & \bf es & \bf fr & \bf it & \bf pt & \bf pl & \bf bg & \bf cs & \bf hu & \bf fi & \bf el & \bf All \\
         \hline
        \multicolumn{18}{l}{\textbf{One-to-one} (Fine-tune \xlmroberta or monolingually pretrained \bert{s} in one language, test in the \emph{same} language.)} \\
        \hline
        \nativebert & \bf 67.7 & 65.5 & \bf 68.4 & 66.7 & \bf 68.5 & \bf 68.5 & 67.6 & \bf 67.4 & \bf  67.9 & \bf 67.4 & \bf 67.2 & - & \bf 66.7 & \bf 67.7 & \bf 67.8 & \bf 67.8 & \bf 67.4 \\
        \xlmroberta & 67.4 & \bf 66.7 & 67.5 & \bf 67.3 & 66.5 & 66.4 & \bf 67.8 & 67.2 & 67.4 & 67.0 & 65.0 & 66.1 & \bf 66.7 & 65.5 & 66.5 & 65.8 & 66.6 \\
        Diff. 		& -0.3 & +1.2 & -0.9 & +0.6 & -2.0 & -2.1 & +0.2 & -0.2 & -0.5 & -0.4 & -2.2 & - & 0.0 & -2.2 & -1.3 & -2.0 & -0.7 \\
        \hline
        \multicolumn{18}{l}{\textbf{One-to-many} (Fine-tune \xlmroberta \emph{only} in English, test in all languages, with alternative adaptation strategies.)} \\
        \hline
        End-to-end fine-tuning  & \bf 67.4 & 56.5 & 52.4 & 49.0 & 55.7 & 55.2 & 54.0 & 55.0 & 52.0 & 50.5 & 46.9 & 51.2 & 49.6 & 48.8 & 46.4 & 33.3 & 49.3 \\
        First 3 blocks frozen & 66.3 & 59.1 & 56.8 & 55.3 & 57.5 & 57.9 & 58.1 & 57.7 & 56.2 & 54.9 & 53.7 & 56.1 & 54.3 & 51.0 & 52.1 & 42.4 & 53.0 \\
        First 6 blocks frozen & 66.3 & 59.1 & 57.4 & 55.7 & 57.9 & 57.2 & 56.9 & 57.9 & 53.9 & 55.4 & 51.9 & 55.8 & 52.6 & 47.3 & 48.7 & 39.6 & 51.7 \\
        First 9 blocks frozen & 65.8 & 59.4 & 57.9 & 56.9 & 58.6 & 58.2 & 58.7 & 59.4 & 55.7 & 57.5 & 53.4 & 56.7 & 54.2 & 48.8 & 50.4 & 44.5 & 53.0 \\
        All 12 blocks frozen & 27.2 & 21.4 & 24.6 & 24.6 & 23.0 & 21.6 & 23.4 & 21.9 & 20.1 & 25.1 & 22.8 & 23.1 & 24.3 & 22.8 & 21.9 & 19.0 & 22.2 \\
        Adapter modules             
        & 67.3 & \bf 61.5 & \bf 59.3 & \bf 57.8 & \bf 59.5 & \bf 60.3 & \bf 61.0 & \bf 60.4 & \bf 58.8 & \bf 58.5 & \bf 57.5 & \bf 59.2 & \bf 56.8 & \bf 55.3 & \bf 55.6 & \bf 46.1 & \bf 56.1 \\
        \bitfit (bias terms only)  & 63.9 & 59.3 & 57.0 & 54.0 & 58.2 & 57.8 & 57.4 & 56.9 & 56.4 & 55.5 & 54.0 & 55.6 & 54.8 & 51.2 & 54.8 & 42.1 & 53.7 \\
        \lnfit (layer-norm only) & 63.1 & 58.9 & 55.7 & 54.1 & 56.6 & 59.1 & 59.1 & 58.0 & 56.6 & 57.2 & 55.7 & 55.4 & 52.8 & 51.4 & 50.7 & 39.9 & 53.3 \\
        \hline
        \multicolumn{18}{l}{\textbf{Many-to-many} (Jointly fine-tune \xlmroberta in \emph{all} languages, test in all languages, with alternative adaptation strategies.)} \\
        \hline
        End-to-end fine-tuning & 66.4 & 66.2 & 66.2 & 66.1 & 66.1 & 66.3 & 66.3 & 66.2 & 66.3 & 65.9 & 65.6 & 65.7 & 65.7 & 65.2 & 65.8 & 65.1 & 65.7 \\
        Adapter modules          & \bf 67.2 & \bf 67.1 & \bf 66.3 & \bf 67.1 & \bf 67.0 & \bf 67.4 & \bf 67.2 & \bf 67.1 & \bf 67.4 & \bf 67.0 & \bf 66.2 & \bf 66.6 & \bf 67.0 & \bf 65.5 & \bf 66.6 & \bf 65.7 & \bf 66.4 \\
    \end{tabular}
    }
    \vspace{-3mm}
    \caption{Test results for level 3 (567 labels) of \multieurlex. We show \mrp (\%) for the 16 most widely spoken \eu official languages, and \mrp averaged over all 23 languages. Appendix~\ref{sec:appendix_a} reports results for all languages.}
    \label{tab:results}
    \vspace{-3mm}
\end{table*}

\noindent\textbf{Evaluation:}  
Given the large number and skewed distribution of labels, retrieval measures have been favored in large-scale multi-label text classification literature \cite{Mullenbach2018, chalkidis2019}. Following \citet{chalkidis2019, chalkidis2020}, we report \emph{mean R-Precision} (\mrp) \cite{manning2009}. That is, for each document, the model ranks the labels it selects by decreasing confidence, 
and we compute Precision@$k$, where $k$ is the document's number of gold labels; we then average over documents. For all experiments, we use the chronological data split and report the average across three runs.
Unless stated otherwise, we use level 3 with $L=567$ labels (Table~\ref{fig:eurovoc}), which has a highly skewed (long-tail) label distribution and temporal concept drift (Table~\ref{tab:kl}). In Section~\ref{sec:cross}, we also consider label sets from the other levels.

\section{Experiments and Discussion}

\vspace*{-2mm}
For the main experiments, we mainly use \xlmroberta in a \emph{one-to-many} setting (fine-tuning in English, testing in all   languages). We also report key \mt results for completeness. As a ceiling for cross-lingual transfer, in Section~\ref{sec:mono} we first evaluate monolingual (native) \bert models and \xlmroberta, both in a \emph{one-to-one} manner (fine-tuning and testing in the same language), which requires annotated training data in the target language. 
For completeness, in Section~\ref{sec:multilingualTraining}, we also report \emph{many-to-many} results, where \xlmroberta is jointly fine-tuned and tested in all languages. 

\subsection{Monolingual Classification (\emph{one-to-one})}
\label{sec:mono}

Table~\ref{tab:results} (top) shows that in the \emph{one-to-one} setting, \xlmroberta is competitive to native (monolingually pretrained) \bert{s} with a minor decrease of 0.7 \mrp on average across languages. Of course, the \emph{one-to-one} setting requires training data in the target language. We report these results as an \emph{upper bound} for zero-shot cross-lingual transfer. Also, the native \bert{s} are pretrained on corpora of different sizes and quality, which explains why they are not consistently better than \xlmroberta.\vspace*{-1mm}

\subsection{Cross-lingual Transfer (\emph{one-to-many})} 
\label{sec:cross}

\textbf{\xlmroberta adaptation:} In the one-to-many setting, where we fine-tune in English and test in all languages, Table~\ref{tab:results} (middle) shows that all adaptation strategies vastly improve the performance of \xlmroberta across languages (up to 6.8 All \mrp increase) comparing to no adaptation (end-to-end fine-tuning), while remaining competitive in English (source). This indicates that not fine-tuning the full set of parameters helps the model retain more of its multilingual knowledge obtained during pretraining. We observe no big difference among the block freezing strategies for $N=$ 3, 6, 9, but performance deteriorates substantially when all blocks are frozen ($N\!=\!$ 12).\footnote{When $N=$ 12, we practically evaluate (probe) the intact pre-training knowledge of \xlmroberta in the end-task.} 
We speculate there is a trade-off between freezing more blocks to retain multilingual knowledge and freezing fewer blocks to benefit end-task (classification) performance. Adapters consistently lead to the best results in all languages and overall (All \mrp 56.1), with practically no decrease in English (source) performance (67.3). \bitfit, which only fine-tunes bias terms (4e-2\% of parameters), and \lnfit, which fine-tunes even fewer parameters (1e-2\%), are the second and third best strategies. These results highlight the 
expressive power of the few parameters \bitfit and \lnfit modify; this observation has been also discussed in previous studies \cite{frankle2021, zaken2021}, but not in a multi-lingual setting. Overall, fine-tuning in a single language leads to substantial forgetting of multilingual knowledge, but adaptation strategies, especially adapter modules, alleviate this problem and improve cross-lingual end-task performance.\vspace{2mm}

\begin{figure*}[!htb]
\centering
    \includegraphics[width=\textwidth]{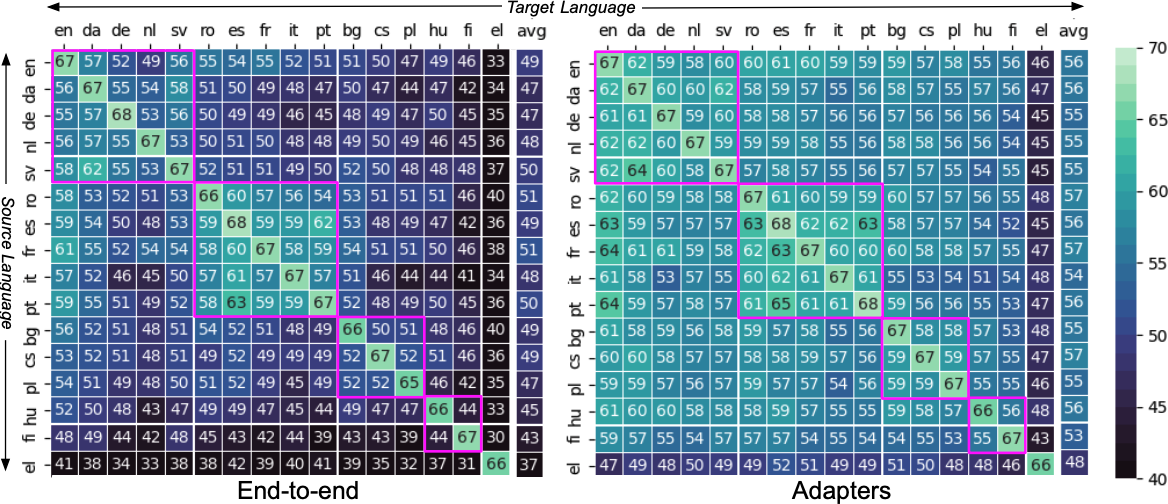}
    \vspace{-7mm}
\caption{Test results (\mrp, \%) for Level 3 (567 labels) with \xlmroberta, when fine-tuning in one language (source, rows) and testing in all languages (columns), without adaptation (end-to-end, left) and with adapter modules (right). The languages are grouped (framed) in language families (Germanic, Romance, Slavic, Uralic).}
\label{fig:heatmaps}
\vspace{-3mm}
\end{figure*}

\noindent\textbf{\mt adaptation:}
In Table~\ref{tab:mt_test}, we repeat the \emph{one-to-many} experiments of Table~\ref{tab:results}, this time with \mt (encoder only).
For brevity, we report only \mrp on the source (English) language, and \mrp averaged over the 23 languages.\footnote{See Appendix~\ref{sec:appendix_a} for additional experimental results.} \bitfit cannot be applied in this case, because \mt does not use bias terms. As in Table~\ref{tab:results}, freezing the initial $N$ blocks of the encoder ($N=3,6,9$) improves cross-lingual transfer (average \mrp increase up to 4.7), but freezing all layers ($N=$ 12) harms performance. Surprisingly adapter modules, which are the best adaptation strategy for \xlmroberta (Table~\ref{tab:results}, middle), lead to very poor performance (average \mrp 44); there are similar results with \lnfit (average \mrp 38.7). We speculate this happens because the encoder of \mt  needs to `re-program' itself during fine-tuning to perform as a stand-alone encoder; in adapters and \lnfit `re-programming' is only facilitated by very few parameters and the model is `forced' (due to low adaptable capacity) to discard multilingual knowledge aggressively. \xlmroberta follows the opposite pattern (fewer parameters lead to better cross-lingual transfer), because it is pre-trained as a stand-alone encoder.
We leave a more thorough investigation of the trade-off between the number of trainable parameters vs. end-task (Src/All) performance for future work.\vspace{2mm}

\begin{table}[t]
    \centering
    \resizebox{\columnwidth}{!}{
    \begin{tabular}{l|rr|c|c}
         \bf Adaptation strategy &  \multicolumn{2}{c|}{\bf Params (\%)} & \bf en (Src) & \bf All \\
         \hline
         End-to-end fine-tuning & 277M & (100.0\%) & 67.4 & 53.7 \\ 
         First 3 blocks frozen & 63.7M & (23.0\%) & 67.4 & 56.9 \\
         First 6 blocks frozen & 42.4M & (15.3\%) & 66.3 & \bf 58.4 \\
         First 9 blocks frozen & 21.2M & (7.7\%) & \bf 68.0 & 58.3\\
         All 12 blocks frozen & -- & (0.0\%) & 20.2 & 16.8 \\
         Adapter modules & 7.1M & (1.7\%) & 66.3 & 44.0 \\
         \lnfit (layer-norm only) & 19.2K & (0.01\%) & 59.5 &  38.7 \\
    \end{tabular}
    }
    \vspace{-2mm}
    \caption{Test results of \mt fine-tuned in English (en). We show \mrp (\%) in English (Src), and averaged across all 23 languages (All). We also report the trainable parameters 
    (excl.\ the classification layer).
    }
    \label{tab:mt_test}
    \vspace{-3mm}
\end{table}

\begin{figure}[t]
    \centering
    \includegraphics[width=\columnwidth]{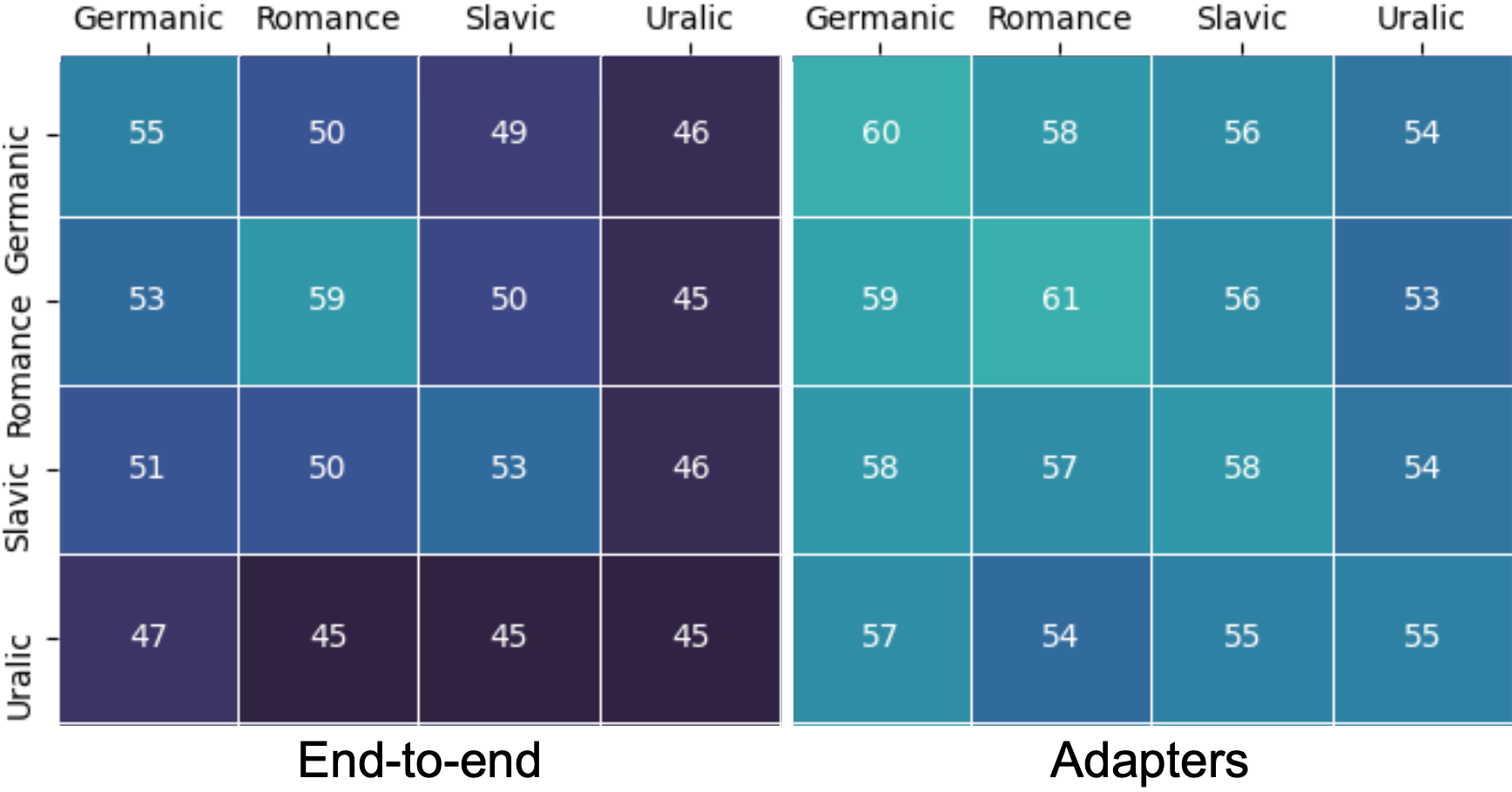}
    \vspace{-8mm}
    \caption{Cross-lingual test results (\mrp, \%) for level 3 (567 labels) with \xlmroberta, averaged 
    over language families (transfer from one family to another).}
    \label{fig:families}
    \vspace{-4mm}
\end{figure}

\begin{figure*}[t]
    \centering
    \resizebox{0.8\textwidth}{!}{
    \includegraphics{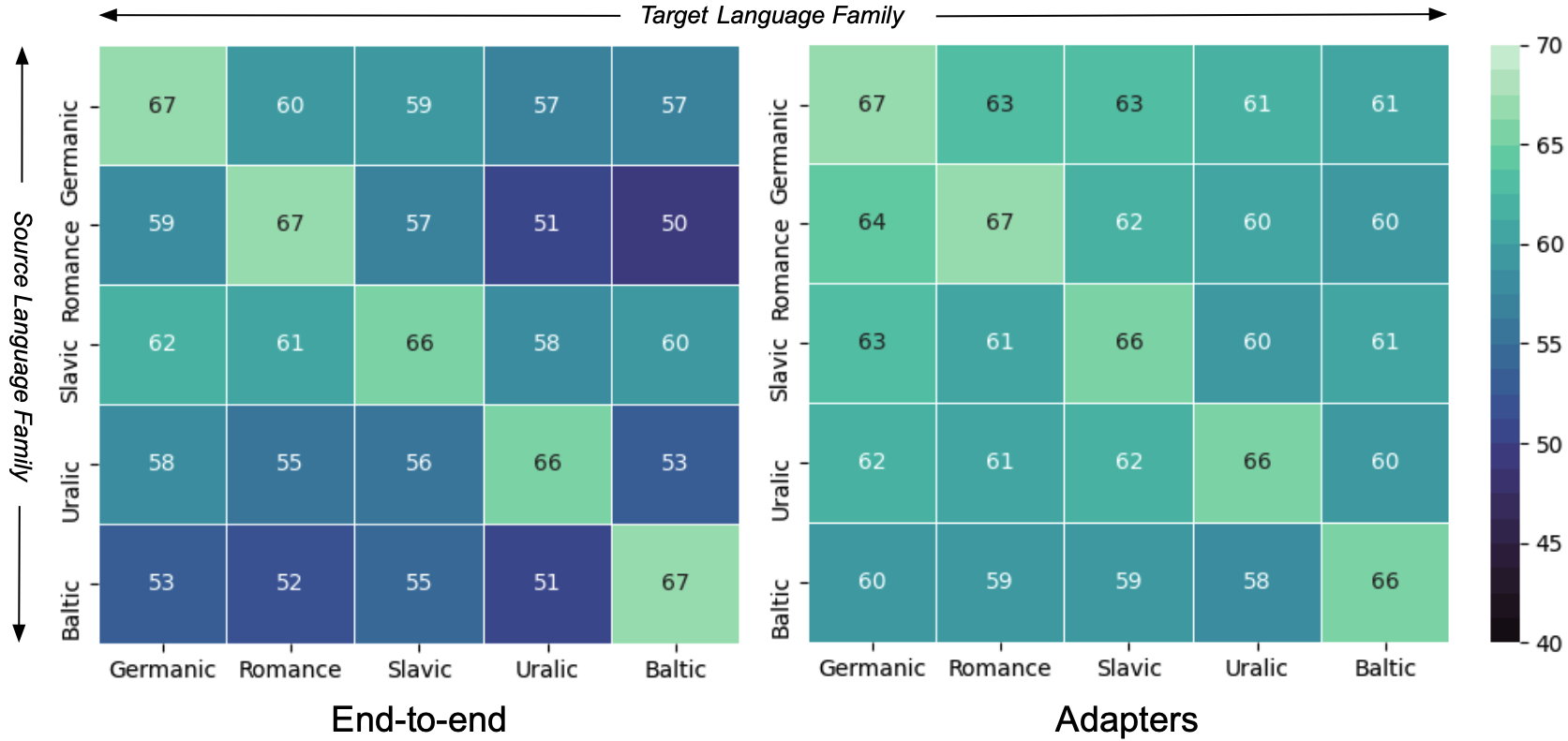}
    }
    \vspace{-1mm}
    \caption{Cross-lingual test results (\mrp, \%) for level 3 (567 labels) with \xlmroberta, when fine-tuning \emph{end-to-end} or with adapters in \emph{all languages of the same family (Src)} and testing is averaged 
    over each language family.}
    \label{fig:family_models}
    \vspace{-2mm}
\end{figure*}

\noindent\textbf{Different source languages:} 
In the cross-lingual experiments so far, we fine-tuned the model in English (source) and evaluated it in all 23 languages. In Fig.~\ref{fig:heatmaps}, we repeat these experiments using a \emph{different source language in each repetition} (rows), evaluating again in all languages (columns).\footnotemark[13]
We use \xlmroberta without adaptation (end-to-end, left) or with adapter modules (right). 
Despite the dominance of English in multi-lingual NLP literature, we observe that using alternative source languages (e.g., Romanian or French) lead to better target results. Similar results have been presented in \citet{turc2021} for other NLP tasks.
As in the previous one-to-many experiments with \xlmroberta (Table~\ref{tab:results}, middle), adapters vastly improve cross-lingual transfer across all cases (e.g., English-en to Danish-da improves from 57 to 62 \mrp), with occasionally slightly lower monolingual performance (e.g., German-de drops from 68 to 67). 
Cross-lingual transfer performs overall better when the source and target languages are in the same family (frames of Fig.~\ref{fig:heatmaps}), especially for Romance languages (Fig.~\ref{fig:families}, diagonal).\footnote{See Appendix~\ref{sec:appendix_a} for additional experimental results.}
Also, when using adapters, cross-lingual performance often drops less abruptly when moving outside of the family of the source language. For example, when fine-tuning in Danish-da, if the test set changes from Swedish-sv to Spanish-es, performance drops from 58 to 50 without adapters (Fig.~\ref{fig:heatmaps}, left), but the change is smoother, from 62 to 59, with adapters (Fig.~\ref{fig:heatmaps}, right). This is better illustrated in the right part of Fig.~\ref{fig:families} (smoother changes across cells per row). These results confirm that adapter modules help retain more multilingual knowledge.\vspace{2mm}

\noindent\textbf{Transfer from one family to another:}
In the previous experiment (``\emph{Different source languages}''), we used a \emph{different source language in each repetition} and evaluated in all languages. To better understand how linguistic proximity between families affects performance, in Figure~\ref{fig:family_models} we present additional experiments in a \emph{many-to-many} setting, where each model is trained across \emph{all languages in the same family} (source) and evaluated across all languages. We use again \xlmroberta without  adaptation  (end-to-end, left) or with adapter modules (right). We observe (Fig.~\ref{fig:family_models}, left) that cross-lingual transfer performs overall better when the source and target families are the same.
Also, when using adapters (Fig.~\ref{fig:family_models}, right), cross-lingual performance drops less abruptly when moving to another family, different from the one (source) whose languages were used for fine-tuning. As expected, the cross-lingual performance of these models (jointly fine-tuned in a language family) is substantially higher than the ones trained in a \emph{one-to-one} setting (Fig.~\ref{fig:heatmaps}--\ref{fig:families}), and closer to that of the models jointly fine-tuned in \emph{all 23 languages} (\emph{many-to-many}, results reported in the lower part of Table~\ref{tab:results}).\vspace{3mm}

\begin{table}[t]
    \centering
    \resizebox{\columnwidth}{!}{
    \begin{tabular}{l|c|c|c|c}
         \multirow{2}{*}{Version of the input text} & \multicolumn{2}{c|}{\textbf{en} (Src)} & \multicolumn{2}{c}{\bf Rest} \\
         & $T$ & \mrp & $T$ (\%) & \mrp \\
         \hline
         Full-text & 100\% & \bf 67.3 & 100\% & \bf 56.1 \\
         w/o digits & 89\% & 67.1 & 88\% & 55.0 \\
         w/o digits \& English vocab. & 22\% & 14.0 & 77\% & 51.5 \\
    \end{tabular}
    }
    \vspace{-2mm}
    \caption{Test results of \xlmroberta (with adapters) removing digits and words used in the English part of \multieurlex during inference. We show \mrp (\%) for English (Src) and averaged over the other 22 languages (Rest). $T$ is the percentage of tokens retained.}
    \vspace*{-4mm}
    \label{tab:input_versions}
\end{table}

\begin{table*}[t]
    \centering
    \resizebox{0.97\textwidth}{!}{
    \begin{tabular}{l|rr|cc||cc||cc||cc}
    \hline
         \multirow{2}{*}{\bf Adaptation Strategy} & \multicolumn{2}{c|}{\multirow{2}{*}{\bf Parameters}} & \multicolumn{2}{c||}{Level 1 (21)} & \multicolumn{2}{c||}{Level 2 (127)} & \multicolumn{2}{c||}{Level 3 (567) } & \multicolumn{2}{c}{Original (7,390)} \\
         \cline{4-11}
           &  & & \textbf{en} (Src) & \bf All & \textbf{en} (Src) & \bf All & \textbf{en} (Src) & \bf All & \textbf{en} (Src) & \bf All \\
         \hline
         End-to-end fine-tuning & 278M & (100\%) & \cellcolor{c83} \bf 83.2 & \cellcolor{c76} 75.7 & \cellcolor{c74} \bf 73.6 & \cellcolor{c59} 58.7  & \cellcolor{c67} \bf 67.4 & \cellcolor{c49} 49.3  & \cellcolor{c48} 47.6 & \cellcolor{c28} \color{white} 27.6 \\
         \hline
         First 3 blocks frozen & 63.8M & (23.0\%) &\cellcolor{c83} 82.9 & \cellcolor{c76} 76.4 & \cellcolor{c71} 71.3 & \cellcolor{c60} 60.2  & \cellcolor{c66} 66.3 & \cellcolor{c53} 53.0  & \cellcolor{c47} 47.3 & \cellcolor{c29} \color{white} 29.0 \\
         First 6 blocks frozen & 42.5M & (15.3\%) & \cellcolor{c82} 82.3 & \cellcolor{c77} 76.7 & \cellcolor{c70} 69.6 & \cellcolor{c61} 61.1  & \cellcolor{c66} 66.3 & \cellcolor{c52} 51.7  & \cellcolor{c47} 47.1 & \cellcolor{c30} \color{white} 30.1 \\
         First 9 blocks frozen & 21.3M & (7.7\%) & \cellcolor{c82} 82.0 & \cellcolor{c75} 74.8 & \cellcolor{c71} 70.7 & \cellcolor{c60} 60.1  & \cellcolor{c66} 65.8 & \cellcolor{c53} 53.0  & \cellcolor{c48} 48.0 & \cellcolor{c33} \color{white}  32.8 \\
         Adapter modules & 9.5M &  (3.3\%) & \cellcolor{c83} 83.1 & \bf \cellcolor{c77} 77.2 & \cellcolor{c72} 72.3 & \cellcolor{c61} \bf 61.2  & \cellcolor{c67} 67.3 & \cellcolor{c56} \bf 56.1  & \cellcolor{c48} 47.9 & \cellcolor{c35} \color{white} \bf 35.1 \\
         \bitfit (bias terms only) & 101K & (0.04\%) &\cellcolor{c82}  82.7 & \cellcolor{c76} 76.1 & \cellcolor{c70} 70.2 & \cellcolor{c60} 60.1  &  \cellcolor{c64} 63.9 &\cellcolor{c54}  53.7  & \cellcolor{c48} \bf 48.3 & \cellcolor{c33} \color{white} 33.9 \\
         \lnfit (layer-norm only) & 36.8K & (0.01\%) & \cellcolor{c82} 81.5 & \cellcolor{c75} 74.9 & \cellcolor{c70} 69.7 & \cellcolor{c59} 59.3  & \cellcolor{c63} 63.1 & \cellcolor{c53} 53.3 & \cellcolor{c43} 43.1 & \cellcolor{c26} \color{white} 26.4 \\
         \multicolumn{1}{c|}{\bf \emph{$\uparrow$ Averaged Adapt. $\uparrow$}} & \multicolumn{2}{c|}{-} & \cellcolor{c82} 82.4 & \cellcolor{c76} 76.0 & \cellcolor{c70} 70.6 & \cellcolor{c60} 60.3 & \cellcolor{c66} 65.5 & \cellcolor{c54} 53.5 & \cellcolor{c47} 47.0 & \cellcolor{c31} \color{white} 31.2 \\
         \hline
         All 12 blocks frozen & - & (0.0\%) & \cellcolor{c61} 61.4 & \cellcolor{c57} 56.5 & \cellcolor{c39} 39.0 & \cellcolor{c32} 31.6 & \cellcolor{c27} \color{white} 27.2 & \cellcolor{c22} \color{white} 22.2 & \cellcolor{c26} \color{white} 26.1 & \cellcolor{c15} \color{white} 15.3 \\

         \hline
    \end{tabular}
    }
    \vspace{-3mm}
    \caption{Test results of \xlmroberta fine-tuned in English, for all adaptation strategies and different  label  granularities  (\eurovoc levels, Table~\ref{tab:eurovoc}). We show \mrp results (\%) for English (Src) and averaged over all 23 languages (All). We also count the trainable parameters, excl.\ the classification layer, which remains the same.}
    \label{tab:adapt_all_levels}
    \vspace{-5mm}
\end{table*}

\noindent\textbf{Removing digits and shared words:} In an ablation study, during inference we remove digits and words that are shared across languages to see to what extent label predictions depend on them. Initially, we eliminate digits, which constitute approx.\ 10\% of the average document length measured in white-space separated tokens. Digits often participate in legal references (e.g., \emph{``established by Regulation No \underline{1468/81}''}) or other coding schemes that may hint \eurovoc concepts (e.g., when specific laws are highly cited). Moreover, inspecting training documents, we observe that vocabulary words (e.g., of Latin origin) are shared to a substantial degree (23\% on average) across languages; thus as a second step we remove approx.\ 25k words used more than 25 times in English documents to break direct cross-lingual alignment. Table~\ref{tab:input_versions} shows that removing digits leads to a small decrease in  \emph{one-to-one} performance (-0.2) and a larger, though still small, decrease in \emph{one-to-many} performance (-1.1). Eliminating shared words (present in the English vocabulary) leads to a further decrease (-3.5) in cross-lingual performance, and English performance of course plunges as the remaining text is very short and severely corrupted.\vspace{3mm}

\noindent\textbf{Different label granularities:}
Table~\ref{tab:adapt_all_levels} shows \xlmroberta results with labels from different \eurovoc levels (Table~\ref{tab:eurovoc}) for all adaptation strategies. As expected, performance deteriorates (approx.\ 5-10\% per level) as the size of the label set increases. Nonetheless, we observe consistent improvements with adaptation strategies compared to full (end-to-end) fine-tuning for all label sets, with the exception of the fully (all 12 blocks) frozen model (last row). Adapters have the best overall performance, but the ranking and impact of the different adaptation strategies varies across levels. Specifically, as the size of the label set increases, the average (Table~\ref{tab:adapt_all_levels}, second-to-last line) adaptation zero-shot (All) performance: (a) improves compared to no adaptation (end-to-end fine-tuning), approx.\ +0.3 $\rightarrow$ +1.6 $\rightarrow$ +4.2 $\rightarrow$ +3.6, as we move from level 1 to the full (original) label set, with a small drop from level 3 to the full label set; and (b) deteriorates more aggressively when comparing it to English (Src) performance, 
 approx.\ -6.4 $\rightarrow$ -10.3 $\rightarrow$ -12.0 $\rightarrow$ -15.8.
The latter (b) is due to the need to model increasingly finer concepts (labels), which complicates cross-lingual concept alignment and, hence, hurts transfer, leaving more scope for adaptation strategies to make a difference (a).\vspace{-2mm}

\subsection{Multilingual Fine-tuning (\emph{many-to-many})} 
\label{sec:multilingualTraining}

In the lower part of Table~\ref{tab:results}, we report results for \xlmroberta, fine-tuned end-to-end or using adapters, when the model is \emph{jointly fine-tuned in all languages}. In this case, for each epoch and batch we randomly select a language of the document, among the available ones; not all documents are available in all 23 languages (Table~\ref{tab:stats}). 
Adapter modules again consistently improve performance. The \emph{many-to-many} models largely outperform the \emph{one-to-many} models (Table~\ref{tab:results}, middle), as they have access to annotated training documents in all languages.
Nevertheless, this is still an interesting scenario, because it allows \emph{deploying a single model} that handles all languages and is competitive to using multiple native \bert models, one per language\vspace{-1mm}

\section{Conclusions and Future Work}
\vspace{-1mm}
We introduced \multieurlex, a new multilingual legal topic classification dataset with 65k documents (\eu laws) in 23 languages, where each document is annotated with multiple labels (concepts) from the \eurovoc taxonomy, with alternative label granularities. To the best of our knowledge, this is one of the most diverse, in terms of languages, classification datasets. We mainly used the dataset as a testbed for zero-shot cross-lingual transfer.

Experimental results showed that fine-tuning a multilingually pretrained model (\xlmroberta, \mt) in a single language leads to catastrophic forgetting of multilingual knowledge and, consequently, poor zero-shot transfer. We found that adaptation strategies, originally proposed to accelerate fine-tuning for end-tasks, help retain multilingual knowledge from pretraining, substantially improving zero-shot cross-lingual transfer. However, their impact depends on the size of the label set, i.e., the gains increase as the label set increases.
Interestingly, even adaptation strategies (\bitfit, \lnfit) that fine-tune a very small fraction of parameters (<0.05\%) are competitive.
Experimental results also showed that multilingual models are competitive to monolingual models in the one-to-one set-up; and that a single multilingual model jointly fine-tuned in all languages is also competitive. We also used \multieurlex to highlight the effect of temporal concept drift and the importance of chronological, instead of random, splits.

In future, we would like to examine alternative cross-lingual adaptation strategies \cite{pfeiffer2020,pfeiffer-etal-2021-adapterfusion} and distributionally robust optimization techniques \cite{sagawa2020,wei2021} to address the temporal concept drift.

\section*{Ethics Statement}

The dataset contains publicly available \eu laws that do not include personal or sensitive information, with the exception of trivial information presented by consent, e.g., the names of the active presidents of the European Parliament, European Council, or other official administration bodies. The collected data is licensed under the Creative Commons Attribution 4.0 International licence (\url{https://eur-lex.europa.eu/content/legal-notice/legal-notice.html}). \multieurlex covers 23 languages from seven language  families (Germanic, Romance, Slavic, Uralic, Baltic, Semitic, Hellenic). This does not imply that no other languages are spoken in \eu countries, although \eu laws are not translated to other languages (\url{https://europa.eu/european-union/about-eu/eu-languages_en}). We also provide a detailed Dataset Card \cite{gebru2018} for the \multieurlex dataset in Appendix~\ref{sec:data_card}.

\section*{Acknowledgments} This work is partly funded by the Innovation Fund Denmark (IFD)\footnote{\url{https://innovationsfonden.dk/en}} under File No.\ 0175-00011A. We would like to thank Prodromos Malakasiotis, Nikolaos Aletras, Anders Søgaard, and Yoav Goldberg for providing valuable feedback, as well as Reviewer \#3 for a particularly thorough review and feedback. We are grateful to Cognitiv+ Ltd.\footnote{\url{https://www.cognitivplus.com}} for providing the compute infrastructure (an \textsc{nvidia dgx}-1 server with 8x \textsc{nvidia V}100 cards) for running the overwhelming number of experiments.

\bibliography{emnlp2021}

\begin{thebibliography}{62}
\expandafter\ifx\csname natexlab\endcsname\relax\def\natexlab#1{#1}\fi

\bibitem[{Aletras et~al.(2019)Aletras, Ash, Barrett, Chen, Meyers,
  Preotiuc-Pietro, Rosenberg, and Stent}]{aletras2019}
Nikolaos Aletras, Elliott Ash, Leslie Barrett, Daniel Chen, Adam Meyers, Daniel
  Preotiuc-Pietro, David Rosenberg, and Amanda Stent, editors. 2019.
\newblock \href {https://www.aclweb.org/anthology/W19-2200} {\emph{Proceedings
  of the Natural Legal Language Processing Workshop 2019}}. Minneapolis,
  Minnesota.

\bibitem[{Aletras et~al.(2016)}]{Aletras2016}
Nikolaos Aletras et~al. 2016.
\newblock Predicting judicial decisions of the {E}uropean {C}ourt of {H}uman
  {R}ights: a {N}atural {L}anguage {P}rocessing perspective.
\newblock \emph{PeerJ Computer Science}, 2:e93.

\bibitem[{Angelidis et~al.(2018)Angelidis, Chalkidis, and
  Koubarakis}]{Angelidis2018}
Iosif Angelidis, Ilias Chalkidis, and Manolis Koubarakis. 2018.
\newblock \href {https://ebooks.iospress.nl/volumearticle/50829} {{Named Entity
  Recognition, Linking and Generation for Greek Legislation}}.
\newblock In \emph{Proceedings of the 31st International Conference on Legal
  Knowledge and Information Systems (JURIX)}, Groningen, The Netherlands.

\bibitem[{Artetxe et~al.(2020)Artetxe, Ruder, and Yogatama}]{artetxe2020}
Mikel Artetxe, Sebastian Ruder, and Dani Yogatama. 2020.
\newblock \href {https://doi.org/10.18653/v1/2020.acl-main.421} {On the
  cross-lingual transferability of monolingual representations}.
\newblock In \emph{Proceedings of the 58th Annual Meeting of the Association
  for Computational Linguistics}, pages 4623--4637, Online.

\bibitem[{Artetxe and Schwenk(2019)}]{Artetxe2019}
Mikel Artetxe and Holger Schwenk. 2019.
\newblock \href {https://doi.org/10.1162/tacl_a_00288} {{Massively Multilingual
  Sentence Embeddings for Zero-Shot Cross-Lingual Transfer and Beyond}}.
\newblock \emph{Transactions of the Association for Computational Linguistics},
  7:597--610.

\bibitem[{Ba et~al.(2016)Ba, Kiros, and Hinton}]{ba2016}
Jimmy~Lei Ba, Jamie~R. Kiros, and Geoffrey~E. Hinton. 2016.
\newblock \href {https://arxiv.org/abs/1607.06450} {Layer normalization}.
\newblock In \emph{NIPS 2016 Deep Learning Symposium}.

\bibitem[{Chalkidis et~al.(2019)Chalkidis, Fergadiotis, Malakasiotis, and
  Androutsopoulos}]{chalkidis2019}
Ilias Chalkidis, Emmanouil Fergadiotis, Prodromos Malakasiotis, and Ion
  Androutsopoulos. 2019.
\newblock \href {https://doi.org/10.18653/v1/P19-1636} {Large-scale multi-label
  text classification on {EU} legislation}.
\newblock In \emph{Proceedings of the 57th Annual Meeting of the Association
  for Computational Linguistics}, pages 6314--6322, Florence, Italy.

\bibitem[{Chalkidis et~al.(2020{\natexlab{a}})Chalkidis, Fergadiotis, Kotitsas,
  Malakasiotis, Aletras, and Androutsopoulos}]{chalkidis2020}
Ilias Chalkidis, Manos Fergadiotis, Sotiris Kotitsas, Prodromos Malakasiotis,
  Nikolaos Aletras, and Ion Androutsopoulos. 2020{\natexlab{a}}.
\newblock \href {https://doi.org/10.18653/v1/2020.emnlp-main.607} {An empirical
  study on large-scale multi-label text classification including few and
  zero-shot labels}.
\newblock In \emph{Proceedings of the 2020 Conference on Empirical Methods in
  Natural Language Processing (EMNLP)}, pages 7503--7515, Online.

\bibitem[{Chalkidis et~al.(2020{\natexlab{b}})Chalkidis, Fergadiotis,
  Malakasiotis, Aletras, and Androutsopoulos}]{chalkidis2020b}
Ilias Chalkidis, Manos Fergadiotis, Prodromos Malakasiotis, Nikolaos Aletras,
  and Ion Androutsopoulos. 2020{\natexlab{b}}.
\newblock \href {https://doi.org/10.18653/v1/2020.findings-emnlp.261}
  {{LEGAL}-{BERT}: The muppets straight out of law school}.
\newblock In \emph{Findings of the Association for Computational Linguistics:
  EMNLP 2020}, pages 2898--2904, Online.

\bibitem[{Chan et~al.(2020)Chan, Schweter, and M{\"o}ller}]{gbert}
Branden Chan, Stefan Schweter, and Timo M{\"o}ller. 2020.
\newblock \href {https://doi.org/10.18653/v1/2020.coling-main.598}
  {{G}erman{'}s next language model}.
\newblock In \emph{Proceedings of the 28th International Conference on
  Computational Linguistics}, pages 6788--6796, Online.

\bibitem[{Conneau et~al.(2020)Conneau, Khandelwal, Goyal, Chaudhary, Wenzek,
  Guzm{\'a}n, Grave, Ott, Zettlemoyer, and Stoyanov}]{conneau2020}
Alexis Conneau, Kartikay Khandelwal, Naman Goyal, Vishrav Chaudhary, Guillaume
  Wenzek, Francisco Guzm{\'a}n, Edouard Grave, Myle Ott, Luke Zettlemoyer, and
  Veselin Stoyanov. 2020.
\newblock \href {https://doi.org/10.18653/v1/2020.acl-main.747} {Unsupervised
  cross-lingual representation learning at scale}.
\newblock In \emph{Proceedings of the 58th Annual Meeting of the Association
  for Computational Linguistics}, pages 8440--8451, Online.

\bibitem[{Conneau et~al.(2018)Conneau, Rinott, Lample, Williams, Bowman,
  Schwenk, and Stoyanov}]{conneau2018}
Alexis Conneau, Ruty Rinott, Guillaume Lample, Adina Williams, Samuel Bowman,
  Holger Schwenk, and Veselin Stoyanov. 2018.
\newblock \href {https://doi.org/10.18653/v1/D18-1269} {{XNLI}: Evaluating
  cross-lingual sentence representations}.
\newblock In \emph{Proceedings of the 2018 Conference on Empirical Methods in
  Natural Language Processing}, pages 2475--2485, Brussels, Belgium.

\bibitem[{Delobelle et~al.(2020)Delobelle, Winters, and
  Berendt}]{delobelle2020robbert}
Pieter Delobelle, Thomas Winters, and Bettina Berendt. 2020.
\newblock \href {https://doi.org/10.18653/v1/2020.findings-emnlp.292}
  {{R}ob{BERT}: a {D}utch {R}o{BERT}a-based {L}anguage {M}odel}.
\newblock In \emph{Findings of the Association for Computational Linguistics:
  EMNLP 2020}, pages 3255--3265, Online. Association for Computational
  Linguistics.

\bibitem[{Devlin et~al.(2019)Devlin, Chang, Lee, and Toutanova}]{devlin2019}
Jacob Devlin, Ming{-}Wei Chang, Kenton Lee, and Kristina Toutanova. 2019.
\newblock \href {https://arxiv.org/abs/1810.04805} {{BERT: Pre-training of Deep
  Bidirectional Transformers for Language Understanding}}.
\newblock In \emph{Proceedings of the Annual Conference of the North American
  Chapter of the Association for Computational Linguistics: Human Language
  Technologies}, volume abs/1810.04805.

\bibitem[{Eisenschlos et~al.(2019)Eisenschlos, Ruder, Czapla, Kadras, Gugger,
  and Howard}]{eisenschlos2019}
Julian Eisenschlos, Sebastian Ruder, Piotr Czapla, Marcin Kadras, Sylvain
  Gugger, and Jeremy Howard. 2019.
\newblock \href {https://doi.org/10.18653/v1/D19-1572} {{M}ulti{F}i{T}:
  Efficient multi-lingual language model fine-tuning}.
\newblock In \emph{Proceedings of the 2019 Conference on Empirical Methods in
  Natural Language Processing and the 9th International Joint Conference on
  Natural Language Processing (EMNLP-IJCNLP)}, pages 5702--5707, Hong Kong,
  China.

\bibitem[{European~Commision(2012)}]{ec2012}
Task force of~the European~Commision. 2012.
\newblock \href
  {https://data.europa.eu/euodp/el/data/dataset/S1049_77_1_EBS386} {{Special
  Eurobarometer 386: Europeans and their Languages}}.
\newblock EU Directorate-General for Communication.

\bibitem[{Frankle et~al.(2021)Frankle, Schwab, and Morcos}]{frankle2021}
Jonathan Frankle, David~J. Schwab, and Ari~S. Morcos. 2021.
\newblock \href {https://arxiv.org/abs/2003.00152} {Training batchnorm and only
  batchnorm: On the expressive power of random features in cnns}.
\newblock In \emph{9th International Conference on Learning Representations
  (ICLR 2021)}, Online.

\bibitem[{Galassi et~al.(2020)Galassi, Drazewski, Lippi, and
  Torroni}]{galassi2020}
Andrea Galassi, Kasper Drazewski, Marco Lippi, and Paolo Torroni. 2020.
\newblock \href {https://doi.org/10.18653/v1/2020.coling-main.79}
  {Cross-lingual annotation projection in legal texts}.
\newblock In \emph{Proceedings of the 28th International Conference on
  Computational Linguistics}, pages 915--926, Barcelona, Spain (Online).

\bibitem[{Gebru et~al.(2018)Gebru, Morgenstern, Vecchione, Vaughan, Wallach,
  III, and Crawford}]{gebru2018}
Timnit Gebru, Jamie Morgenstern, Briana Vecchione, Jennifer~Wortman Vaughan,
  Hanna~M. Wallach, Hal~Daum{\'{e}} III, and Kate Crawford. 2018.
\newblock \href {http://arxiv.org/abs/1803.09010} {Datasheets for datasets}.
\newblock \emph{CoRR}, abs/1803.09010.

\bibitem[{Gonalves and Quaresma(2010)}]{gonalves2010}
Teresa Gonalves and Paulo Quaresma. 2010.
\newblock \href {http://www.di.uevora.pt/~tcg/papers/tcg10b-multilingual.pdf}
  {Multilingual text classification through combination of monolingual
  classifiers}.
\newblock In \emph{CEUR Workshop}, volume 605.

\bibitem[{Gorman and Bedrick(2019)}]{gorman-bedrick-2019-need}
Kyle Gorman and Steven Bedrick. 2019.
\newblock \href {https://www.aclweb.org/anthology/P19-1267} {We need to talk
  about standard splits}.
\newblock In \emph{Proceedings of the 57th Annual Meeting of the Association
  for Computational Linguistics}, pages 2786--2791, Florence, Italy.

\bibitem[{Guti{\'{e}}rrez{-}Fandi{\~{n}}o
  et~al.(2021)Guti{\'{e}}rrez{-}Fandi{\~{n}}o, Armengol{-}Estap{\'{e}},
  P{\`{a}}mies, Llop{-}Palao, Silveira{-}Ocampo, Carrino, Gonzalez{-}Agirre,
  Armentano{-}Oller, Penagos, and Villegas}]{spanishroberta}
Asier Guti{\'{e}}rrez{-}Fandi{\~{n}}o, Jordi Armengol{-}Estap{\'{e}}, Marc
  P{\`{a}}mies, Joan Llop{-}Palao, Joaqu{\'{\i}}n Silveira{-}Ocampo,
  Casimiro~Pio Carrino, Aitor Gonzalez{-}Agirre, Carme Armentano{-}Oller,
  Carlos~Rodr{\'{\i}}guez Penagos, and Marta Villegas. 2021.
\newblock \href {http://arxiv.org/abs/2107.07253} {Spanish language models}.
\newblock \emph{CoRR}, abs/2107.07253.

\bibitem[{He et~al.(2016)He, Zhang, Ren, and Sun}]{he2016}
Kaiming He, Xiangyu Zhang, Shaoqing Ren, and Jian Sun. 2016.
\newblock \href {https://doi.org/10.1109/CVPR.2016.90} {Deep residual learning
  for image recognition}.
\newblock In \emph{2016 IEEE Conference on Computer Vision and Pattern
  Recognition (CVPR)}, pages 770--778, Las Vegas, NV, USA.

\bibitem[{Hendrycks et~al.(2021)Hendrycks, Burns, Chen, and
  Ball}]{hendrycks2021cuad}
Dan Hendrycks, Collin Burns, Anya Chen, and Spencer Ball. 2021.
\newblock \href {https://arxiv.org/abs/2103.06268} {Cuad: An expert-annotated
  nlp dataset for legal contract review}.
\newblock \emph{arXiv preprint arXiv:2103.06268}.

\bibitem[{Houlsby et~al.(2019)Houlsby, Giurgiu, Jastrzebski, Morrone,
  de~Laroussilhe, Gesmundo, Attariyan, and Gelly}]{houlsby2019}
Neil Houlsby, Andrei Giurgiu, Stanislaw Jastrzebski, Bruna Morrone, Quentin
  de~Laroussilhe, Andrea Gesmundo, Mona Attariyan, and Sylvain Gelly. 2019.
\newblock \href {https://arxiv.org/abs/1902.00751} {Parameter-efficient
  transfer learning for nlp}.
\newblock In \emph{Proceedings of the 36th International Conference on Machine
  Learning (ICML)}, Long Beach, CA, USA.

\bibitem[{Hu et~al.(2020)Hu, Ruder, Siddhant, Neubig, Firat, and
  Johnson}]{hu2020}
Junjie Hu, Sebastian Ruder, Aditya Siddhant, Graham Neubig, Orhan Firat, and
  Melvin Johnson. 2020.
\newblock \href {http://proceedings.mlr.press/v119/hu20b.html} {{XTREME}: A
  massively multilingual multi-task benchmark for evaluating cross-lingual
  generalisation}.
\newblock In \emph{Proceedings of the 37th International Conference on Machine
  Learning}, volume 119 of \emph{Proceedings of Machine Learning Research},
  pages 4411--4421.

\bibitem[{Huang and Paul(2019)}]{huang2019}
Xiaolei Huang and Michael~J. Paul. 2019.
\newblock \href {https://doi.org/10.18653/v1/P19-1403} {Neural temporality
  adaptation for document classification: Diachronic word embeddings and domain
  adaptation models}.
\newblock In \emph{Proceedings of the 57th Annual Meeting of the Association
  for Computational Linguistics}, pages 4113--4123, Florence, Italy.
  Association for Computational Linguistics.

\bibitem[{Ioffe and Szegedy(2015)}]{ioffe2015}
Sergey Ioffe and Christian Szegedy. 2015.
\newblock \href {http://arxiv.org/abs/1502.03167} {Batch normalization:
  Accelerating deep network training by reducing internal covariate shift}.
\newblock \emph{CoRR}, abs/1502.03167.

\bibitem[{Kim et~al.(2015)Kim, Xu, and Goebel}]{Kim2015}
Mi-young Kim, Ying Xu, and Randy Goebel. 2015.
\newblock \href
  {https://webdocs.cs.ualberta.ca/~miyoung2/Papers/JURISIN2015.pdf} {{A
  Convolutional Neural Network in Legal Question Answering}}.
\newblock \emph{Ninth International Workshop on Juris-informatics (JURISIN)}.

\bibitem[{Kingma and Ba(2015)}]{Kingma2015}
Diederik~P. Kingma and Jim Ba. 2015.
\newblock \href {https://arxiv.org/abs/1412.6980} {Adam: A method for
  stochastic optimization}.
\newblock In \emph{Proceedings of the International Conference on Learning
  Representations (ICLR)}, San Diego, CA, USA.

\bibitem[{Koh et~al.(2021)Koh, Sagawa, Marklund, Xie, Zhang, Balsubramani, Hu,
  Yasunaga, Phillips, Beery, Leskovec, Kundaje, Pierson, Levine, Finn, and
  Liang}]{wei2021}
Pang~Wei Koh, Shiori Sagawa, Henrik Marklund, Sang~Michael Xie, Marvin Zhang,
  Akshay Balsubramani, Weihua Hu, Michihiro Yasunaga, Richard~Lanas Phillips,
  Sara Beery, Jure Leskovec, Anshul Kundaje, Emma Pierson, Sergey Levine,
  Chelsea Finn, and Percy Liang. 2021.
\newblock \href {http://arxiv.org/abs/2012.07421} {{WILDS:} {A} benchmark of
  in-the-wild distribution shifts}.
\newblock \emph{CoRR}, abs/2012.07421.

\bibitem[{Koutsikakis et~al.(2020)Koutsikakis, Chalkidis, Malakasiotis, and
  Androutsopoulos}]{greekbert}
John Koutsikakis, Ilias Chalkidis, Prodromos Malakasiotis, and Ion
  Androutsopoulos. 2020.
\newblock \href {https://doi.org/10.1145/3411408.3411440} {Greek-bert: The
  greeks visiting sesame street}.
\newblock In \emph{11th Hellenic Conference on Artificial Intelligence}, SETN
  2020, page 110–117, New York, NY, USA. Association for Computing Machinery.

\bibitem[{Lazaridou et~al.(2021)Lazaridou, Kuncoro, Gribovskaya, Agrawal,
  Liska, Terzi, Gimenez, de~Masson~d'Autume, Ruder, Yogatama, Cao,
  Kocisk{\'{y}}, Young, and Blunsom}]{lazaridou2021}
Angeliki Lazaridou, Adhiguna Kuncoro, Elena Gribovskaya, Devang Agrawal, Adam
  Liska, Tayfun Terzi, Mai Gimenez, Cyprien de~Masson~d'Autume, Sebastian
  Ruder, Dani Yogatama, Kris Cao, Tom{\'{a}}s Kocisk{\'{y}}, Susannah Young,
  and Phil Blunsom. 2021.
\newblock \href {http://arxiv.org/abs/2102.01951} {Pitfalls of static language
  modelling}.
\newblock \emph{CoRR}, abs/2102.01951.

\bibitem[{Liu et~al.(2020)Liu, Gu, Goyal, Li, Edunov, Ghazvininejad, Lewis, and
  Zettlemoyer}]{liu2020}
Yinhan Liu, Jiatao Gu, Naman Goyal, Xian Li, Sergey Edunov, Marjan
  Ghazvininejad, Mike Lewis, and Luke Zettlemoyer. 2020.
\newblock \href {http://arxiv.org/abs/2001.08210} {Multilingual denoising
  pre-training for neural machine translation}.

\bibitem[{Liu et~al.(2019)Liu, Ott, Goyal, Du, Joshi, Chen, Levy, Lewis,
  Zettlemoyer, and Stoyanov}]{liu2019}
Yinhan Liu, Myle Ott, Naman Goyal, Jingfei Du, Mandar Joshi, Danqi Chen, Omer
  Levy, Mike Lewis, Luke Zettlemoyer, and Veselin Stoyanov. 2019.
\newblock \href {http://arxiv.org/abs/1907.11692} {Roberta: {A} robustly
  optimized {BERT} pretraining approach}.
\newblock \emph{CoRR}, abs/1907.11692.

\bibitem[{Luz~de Araujo et~al.(2020)Luz~de Araujo, de~Campos, Ataides~Braz, and
  Correia~da Silva}]{dearaujo2020}
Pedro~Henrique Luz~de Araujo, Te{\'o}filo~Em{\'\i}dio de~Campos, Fabricio
  Ataides~Braz, and Nilton Correia~da Silva. 2020.
\newblock \href {https://www.aclweb.org/anthology/2020.lrec-1.181} {{VICTOR}: a
  dataset for {B}razilian legal documents classification}.
\newblock In \emph{Proceedings of the 12th Language Resources and Evaluation
  Conference}, pages 1449--1458, Marseille, France.

\bibitem[{Manning et~al.(2009)Manning, Raghavan, and Schütze}]{manning2009}
Christopher~D. Manning, Prabhakar Raghavan, and Hinrich Schütze. 2009.
\newblock \href {https://nlp.stanford.edu/IR-book} {\emph{{Introduction to
  Information Retrieval}}}.
\newblock Cambridge University Press.

\bibitem[{Martin et~al.(2020)Martin, Muller, Ortiz~Su{\'a}rez, Dupont, Romary,
  de~la Clergerie, Seddah, and Sagot}]{camembert}
Louis Martin, Benjamin Muller, Pedro~Javier Ortiz~Su{\'a}rez, Yoann Dupont,
  Laurent Romary, {\'E}ric de~la Clergerie, Djam{\'e} Seddah, and Beno{\^\i}t
  Sagot. 2020.
\newblock \href {https://doi.org/10.18653/v1/2020.acl-main.645} {{C}amem{BERT}:
  a tasty {F}rench language model}.
\newblock In \emph{Proceedings of the 58th Annual Meeting of the Association
  for Computational Linguistics}, pages 7203--7219, Online.

\bibitem[{Mencia and Fürnkranz(2007)}]{Mencia2007}
Eneldo~Loza Mencia and Johannes Fürnkranz. 2007.
\newblock \href
  {https://link.springer.com/chapter/10.1007/978-3-642-12837-0_11} {{Efficient
  Multilabel Classification Algorithms for Large-Scale Problems in the Legal
  Domain}}.
\newblock In \emph{Proceedings of the LWA 2007}, pages 126--132.

\bibitem[{Mullenbach et~al.(2018)Mullenbach, Wiegreffe, Duke, Sun, and
  Eisenstein}]{Mullenbach2018}
James Mullenbach, Sarah Wiegreffe, Jon Duke, Jimeng Sun, and Jacob Eisenstein.
  2018.
\newblock \href {http://www.aclweb.org/anthology/N18-1100} {{Explainable
  Prediction of Medical Codes from Clinical Text}}.
\newblock In \emph{Proceedings of the 2018 Conference of the North American
  Chapter of the Association for Computational Linguistics: Human Language
  Technologies}, pages 1101--1111.

\bibitem[{Nemeskey(2020)}]{hubert}
Dávid~Márk Nemeskey. 2020.
\newblock \href {https://doktori.hu/index.php?menuid=193&lang=EN&vid=22732}
  {\emph{Natural Language Processing Methods for Language Modeling}}.
\newblock Ph.D. thesis, E\"otv\"os Lor\'and University.

\bibitem[{Nguyen et~al.(2018)Nguyen, Nguyen, Tojo, Satoh, and
  Shimazu}]{Truong2017}
Son Nguyen, Le-Minh Nguyen, Satoshi Tojo, Ken Satoh, and Akira Shimazu. 2018.
\newblock \href {https://doi.org/10.1007/s10506-018-9225-1} {Recurrent neural
  network-based models for recognizing requisite and effectuation parts in
  legal texts}.
\newblock \emph{Artificial Intelligence and Law}, 26:1--31.

\bibitem[{Pfeiffer et~al.(2021)Pfeiffer, Kamath, R{\"u}ckl{\'e}, Cho, and
  Gurevych}]{pfeiffer-etal-2021-adapterfusion}
Jonas Pfeiffer, Aishwarya Kamath, Andreas R{\"u}ckl{\'e}, Kyunghyun Cho, and
  Iryna Gurevych. 2021.
\newblock \href {https://aclanthology.org/2021.eacl-main.39}
  {{A}dapter{F}usion: Non-destructive task composition for transfer learning}.
\newblock In \emph{Proceedings of the 16th Conference of the European Chapter
  of the Association for Computational Linguistics: Main Volume}, pages
  487--503, Online. Association for Computational Linguistics.

\bibitem[{Pfeiffer et~al.(2020)Pfeiffer, Vuli{\'c}, Gurevych, and
  Ruder}]{pfeiffer2020}
Jonas Pfeiffer, Ivan Vuli{\'c}, Iryna Gurevych, and Sebastian Ruder. 2020.
\newblock \href {https://doi.org/10.18653/v1/2020.emnlp-main.617} {{MAD-X}:
  {A}n {A}dapter-{B}ased {F}ramework for {M}ulti-{T}ask {C}ross-{L}ingual
  {T}ransfer}.
\newblock In \emph{Proceedings of the 2020 Conference on Empirical Methods in
  Natural Language Processing (EMNLP)}, pages 7654--7673, Online.

\bibitem[{Raffel et~al.(2020)Raffel, Shazeer, Roberts, Lee, Narang, Matena,
  Zhou, Li, and Liu}]{raffel2020}
Colin Raffel, Noam Shazeer, Adam Roberts, Katherine Lee, Sharan Narang, Michael
  Matena, Yanqi Zhou, Wei Li, and Peter~J. Liu. 2020.
\newblock \href {http://jmlr.org/papers/v21/20-074.html} {Exploring the limits
  of transfer learning with a unified text-to-text transformer}.
\newblock \emph{Journal of Machine Learning Research}, 21(140):1--67.

\bibitem[{Rosenfeld and Tsotsos(2019)}]{rosenfeld2018}
Amir Rosenfeld and John~K. Tsotsos. 2019.
\newblock \href {http://arxiv.org/abs/1802.00844} {Intriguing properties of
  randomly weighted networks: Generalizing while learning next to nothing}.
\newblock In \emph{16th Conference on Computer and Robot Vision (CRV 2021)},
  Kingston, ON, Canada.

\bibitem[{Ruder et~al.(2021)Ruder, Constant, Botha, Siddhant, Firat, Fu, Liu,
  Hu, Neubig, and Johnson}]{ruder2021}
Sebastian Ruder, Noah Constant, Jan Botha, Aditya Siddhant, Orhan Firat, Jinlan
  Fu, Pengfei Liu, Junjie Hu, Graham Neubig, and Melvin Johnson. 2021.
\newblock \href {http://arxiv.org/abs/2104.07412} {{XTREME-R:} towards more
  challenging and nuanced multilingual evaluation}.
\newblock \emph{CoRR}, abs/2104.07412.

\bibitem[{Ruder et~al.(2017)Ruder, Vulić, and Søgaard}]{ruder2017}
Sebastian Ruder, Ivan Vulić, and Anders Søgaard. 2017.
\newblock \href {http://arxiv.org/abs/1706.04902} {A survey of cross-lingual
  embedding models}.
\newblock \emph{CoRR}, abs/1706.04902.

\bibitem[{Sagawa et~al.(2020)Sagawa, Koh, Hashimoto, and Liang}]{sagawa2020}
Shiori Sagawa, Pang~Wei Koh, Tatsunori~B. Hashimoto, and Percy Liang. 2020.
\newblock \href {http://arxiv.org/abs/1911.08731} {Distributionally robust
  neural networks for group shifts: On the importance of regularization for
  worst-case generalization}.
\newblock In \emph{8th International Conference on Learning Representations
  (ICLR 2020)}.

\bibitem[{Sido et~al.(2021)Sido, Pražák, Přibáň, Pašek, Seják, and
  Konopík}]{czert}
Jakub Sido, Ondřej Pražák, Pavel Přibáň, Jan Pašek, Michal Seják, and
  Miloslav Konopík. 2021.
\newblock \href {http://arxiv.org/abs/2103.13031} {Czert -- czech bert-like
  model for language representation}.
\newblock \emph{arXiv preprint arXiv:2103.13031}.

\bibitem[{Souza et~al.(2020)Souza, Nogueira, and Lotufo}]{bertimbau}
F{\'a}bio Souza, Rodrigo Nogueira, and Roberto Lotufo. 2020.
\newblock \href
  {https://link.springer.com/chapter/10.1007/978-3-030-61377-8_28} {Bertimbau:
  Pretrained bert models for brazilian portuguese}.
\newblock In \emph{Intelligent Systems}, pages 403--417, Cham. Springer
  International Publishing.

\bibitem[{Stahlberg(2020)}]{stahlberg2020}
Felix Stahlberg. 2020.
\newblock \href {http://arxiv.org/abs/1912.02047} {Neural machine translation:
  {A} review}.
\newblock \emph{Journal of Artificial Intelligence Research}, 69.

\bibitem[{Szegedy et~al.(2016)Szegedy, Vanhoucke, Ioffe, Shlens, and
  Wojna}]{szegedy2016}
Christian Szegedy, Vincent Vanhoucke, Sergey Ioffe, Jonathon Shlens, and
  Zbigniew Wojna. 2016.
\newblock \href {http://arxiv.org/abs/1512.00567} {Rethinking the inception
  architecture for computer vision}.
\newblock In \emph{IEEE conference on computer vision and pattern recognition}.

\bibitem[{Søgaard et~al.(2021)Søgaard, Ebert, Bastings, and
  Filippova}]{sogaard2021}
Anders Søgaard, Sebastian Ebert, Jasmijn Bastings, and Katja Filippova. 2021.
\newblock \href {https://www.aclweb.org/anthology/2020.emnlp-main.617} {We need
  to talk about random splits}.
\newblock In \emph{Proceedings of the 2021 Conference of the European Chapter
  of the Association for Computational Linguistics (EACL)}, Online.

\bibitem[{Turc et~al.(2021)Turc, Lee, Eisenstein, Chang, and
  Toutanova}]{turc2021}
Iulia Turc, Kenton Lee, Jacob Eisenstein, Ming{-}Wei Chang, and Kristina
  Toutanova. 2021.
\newblock \href {http://arxiv.org/abs/2106.16171} {Revisiting the primacy of
  english in zero-shot cross-lingual transfer}.
\newblock \emph{CoRR}, abs/2106.16171.

\bibitem[{Vaswani et~al.(2017)Vaswani, Shazeer, Parmar, Uszkoreit, Jones,
  Gomez, Kaiser, and Polosukhin}]{vaswani2017}
Ashish Vaswani, Noam Shazeer, Niki Parmar, Jakob Uszkoreit, Llion Jones,
  Aidan~N. Gomez, Lukasz Kaiser, and Illia Polosukhin. 2017.
\newblock \href
  {https://papers.nips.cc/paper/7181-attention-is-all-you-need.pdf} {{Attention
  Is All You Need}}.
\newblock In \emph{31th Annual Conference on Neural Information Processing
  Systems}, USA.

\bibitem[{Virtanen et~al.(2019)Virtanen, Kanerva, Ilo, Luoma, Luotolahti,
  Salakoski, Ginter, and Pyysalo}]{finbert}
Antti Virtanen, Jenna Kanerva, Rami Ilo, Jouni Luoma, Juhani Luotolahti, Tapio
  Salakoski, Filip Ginter, and Sampo Pyysalo. 2019.
\newblock \href {http://arxiv.org/abs/1912.07076} {Multilingual is not enough:
  {BERT} for finnish}.
\newblock \emph{CoRR}, abs/1912.07076.

\bibitem[{Waltl et~al.(2017)Waltl, Muhr, Glaser, Bonczek, Scepankova, and
  Matthes}]{Waltl2017}
Bernhard Waltl, Johannes Muhr, Ingo Glaser, Georg Bonczek, Elena Scepankova,
  and Florian Matthes. 2017.
\newblock \href {https://ebooks.iospress.nl/volumearticle/48039} {Classifying
  legal norms with active machine learning}.
\newblock In \emph{30th International Conference on Legal Knowledge and
  Information Systems (JURIX)}, pages 11--20, Luxembourg.

\bibitem[{Wang et~al.(2018)Wang, Singh, Michael, Hill, Levy, and
  Bowman}]{wang2018}
Alex Wang, Amanpreet Singh, Julian Michael, Felix Hill, Omer Levy, and Samuel
  Bowman. 2018.
\newblock \href {https://doi.org/10.18653/v1/W18-5446} {{GLUE}: A multi-task
  benchmark and analysis platform for natural language understanding}.
\newblock In \emph{Proceedings of the 2018 {EMNLP} Workshop {B}lackbox{NLP}:
  Analyzing and Interpreting Neural Networks for {NLP}}, pages 353--355,
  Brussels, Belgium.

\bibitem[{Xue et~al.(2021)Xue, Constant, Roberts, Kale, Al-Rfou, Siddhant,
  Barua, and Raffel}]{xue2021}
Linting Xue, Noah Constant, Adam Roberts, Mihir Kale, Rami Al-Rfou, Aditya
  Siddhant, Aditya Barua, and Colin Raffel. 2021.
\newblock \href {https://doi.org/10.18653/v1/2021.naacl-main.41} {m{T}5: A
  massively multilingual pre-trained text-to-text transformer}.
\newblock In \emph{Proceedings of the 2021 Conference of the North American
  Chapter of the Association for Computational Linguistics: Human Language
  Technologies}, pages 483--498, Online. Association for Computational
  Linguistics.

\bibitem[{Zaken et~al.(2021)Zaken, Ravfogel, and Goldberg}]{zaken2021}
Elad~Ben Zaken, Shauli Ravfogel, and Yoav Goldberg. 2021.
\newblock \href {http://arxiv.org/abs/2106.10199} {Bitfit: Simple
  parameter-efficient fine-tuning for transformer-based masked
  language-models}.
\newblock \emph{CoRR}, abs/2106.10199.

\bibitem[{Zhong et~al.(2020)Zhong, Xiao, Tu, Zhang, Liu, and Sun}]{zhong2020}
Haoxi Zhong, Chaojun Xiao, Cunchao Tu, Tianyang Zhang, Zhiyuan Liu, and Maosong
  Sun. 2020.
\newblock \href {https://doi.org/10.18653/v1/2020.acl-main.466} {How does {NLP}
  benefit legal system: A summary of legal artificial intelligence}.
\newblock In \emph{Proceedings of the 58th Annual Meeting of the Association
  for Computational Linguistics}, pages 5218--5230, Online.

\end{thebibliography}
\bibliographystyle{acl_natbib}

\appendix

\section{Implementation Details}
\label{sec:appendix_e}

\subsection{Hyper-parameter Tuning}
Similarly to previous work with pretrained \transformer{-based} models, we conduct grid-search to find the optimal learning rate per method considering classification performance on development data. We use early stopping on development data, if there is no improvement of \mrp for five epochs.
For end-to-end and partial (first $N$ blocks frozen) fine-tuning, we search in \{4e-5, 3e-5, 2e-5, 1e-5\}, as suggested by \citet{devlin2019}; we also include in the search an even smaller learning rate (1e-6) as proposed by \citet{conneau2020}. For all (native) \bert models and \xlmroberta, 3e-5 provided the best development results. For \mt we search in \{1e-3, 1e-4, 3e-5\}, while \citet{xue2021} proposed a fixed learning rate of 1e-3; in our case, 1e-4 provided the best development results. When we use adapter modules, \bitfit, or \lnfit, we search in \{1e-3, 1e-4, 3e-5\}, following \citet{houlsby2019}; again 1e-4 gave the best development results. While \citet{houlsby2019} reported stable results across learning rates, in our case 1e-3 led to very unstable training with terrible performance. 

For the bottleneck in adapter modules, where we have to select the number of hidden units ($K$), we search in \{64, 128, 256, 384, 512\}; 256 gave us the best development results, while the rest are comparable (Table~\ref{tab:adapt_k}). 

\begin{table}[h]
    \centering
\resizebox{0.6\columnwidth{}}{!}{
    \begin{tabular}{c|c|c|c}
         $K$ & Params & \textbf{en} (Src) & \bf All \\
         \hline
         64 & 2.4M & 72.5 & 57.8 \\
         128 & 4.8M & 72.9 & 59.6 \\ 
         256 & 9.5M & 72.5 & 60.2 \\ 
         384 & 14.2M & 73.5 & 58.7 \\ 
         512 & 18.9M & 72.6 & 56.6 \\ 
    \end{tabular}
}
\vspace{-2mm}
    \caption{Development results for different values of $K$ in adapter modules. We show \mrp results (\%) on English development data (Src), and development \mrp averaged over all 23 languages (All). We also report the number of trainable parameters (in millions).}
    \vspace*{-4mm}
    \label{tab:adapt_k}
\end{table}

\subsection{Other Technical Details} 
Given the large length of the documents (450 tokens on average), presented in Table~\ref{tab:stats}, we truncate the documents, if needed, and use the first 512 tokens (sub-words) across all methods. \citet{chalkidis2019} experimented with \rnn{-based} methods using the full or truncated (up to 512 tokens) documents of \textsc{eurlex57k} (English only, 7.4k labels), reporting almost identical results, i.e., the first 512 tokens of a document are adequate. 

We also use label smoothing \cite{szegedy2016} ($\alpha\!=\!0.2$) for levels 1--3, as we found it improves cross-lingual transfer in preliminary experiments. Label smoothing severely harms performance in experiments with the original label assignment (full label set with 7.4k labels).

All experiments ran on an \textsc{nvidia dgx-1} station with 8 \textsc{nvidia v100 16gb gpu} cards, although each experiment (model) was running on a single \textsc{gpu} card at a time. In Table~\ref{tab:times}, we report the average run-time per training experiment.

\begin{table}[h]
    \centering
\resizebox{\columnwidth}{!}{
    \begin{tabular}{l|c|c}
         Model (Strategy) & $L_{train}$ & Avg. run-time \\
         \hline 
         \nativebert (Full) & 1 & 10h \\
         \xlmroberta (Full) & 1 & 12h \\ 
         \xlmroberta (Full) & 23 & 12h \\
         \hline
         >> First 3 blocks frozen & 1 & 11h \\
         >> First 6 blocks frozen & 1 & 8h \\
         >> First 9 blocks frozen & 1 & 7h \\
         >> All 12 blocks frozen  & 1 & 3h \\
         >> Adapter modules & 1 & 10h \\
         >> Adapter modules & 23 & 15h \\
         >> \bitfit & 1 & 18h \\
         >> \lnfit & 1 & 11h \\

    \end{tabular}
    }
    \caption{Average training run-time across methods for Level 3 (575). $L_{train}$ is the number of training languages. While \bitfit and \lnfit only tune a very small fraction (approx.\ 1-4$\times$ 1e-3\%) of the parameters, training takes equal or longer, because the models are trained for more epochs and also there are trainable parameters as low as in the first \transformer block.}
    \vspace*{-4mm}
    \label{tab:times}
\end{table}

\section{Decoder Variants of \mt}
\label{sec:appendix_mt}

\textbf{generative}:
In preliminary experiments, we experimented with \mt's generative fine-tuning, using both the encoder and the decoder, as proposed by \citet{xue2021}. First, we ordered alphabetically the labels ($l_1, l_2, \dots, l_N$) by their identifiers. 
At each timestep, the decoder generates a token representing a label; i.e., we generate \texttt{\small [id\_1]} for $l_1$, \texttt{\small [id\_2]} for $l_2$, etc. 
Similarly to \mt v1.1, we use a new (randomly initialized) classification layer (for a fixed vocabulary representing the $N$ labels) to generate the output token at each timestep, based on the hidden state of the decoder.
The entire model is trained to predict the labels in alphabetic order (in terms of \texttt{\small [id\_i]}, where $i = 1,2, \dots, N$), but we ignore the order of the generated (predicted) labels during evaluation, to not penalize the model for not respecting the order. To rank the predicted labels when computing \mrp, we use the probabilities (over the output vocabulary) assigned by the decoder to the corresponding generated tokens. We call \emph{generative} this (original) version of \mt. 

\smallskip\noindent\textbf{decode-cls}:
We also examined  another \mt variant, where again both the encoder and the decoder are used. In this variant, \emph{decode-cls}, we feed the decoder with a single \cls token (only one decoding timestep); by contrast, in \emph{generative} the decoder performs multiple timesteps, and at each one it is fed with the output generated so far (or the corresponding gold output up to the previous timestep during training). 
In effect, in its single timestep, the decoder of \emph{decode-cls} iteratively (at each decoder block) performs cross-attention over the encoder's output, using an updated query (\cls representation). 
We pass the final representation of the decoder's \cls to the same classification layer we use in the encoder-only variant of \mt (Section~\ref{sec:methods}). 
Both \emph{decode-cls} and \emph{generative} use 12 encoder and 12 decoder blocks, 391M parameters.

\smallskip\noindent\textbf{first-pool, last-pool:}
Finally, we examine another encoder-only variant of \mt, \emph{last-pool}, in addition to the encoder-only variant of Section~\ref{sec:methods}, which we now call \emph{first-pool} to highlight the difference between them. In \emph{last-pool}, we use the encoder's top-level representation of the \s special token of \mt, which is always at the end of the input, to represent the document. Since the position of \s is not always the same, its representation is also affected by its positional embedding. By contrast, in \emph{first-pool} the \cls token is always first, hence its positional embedding does not vary. 

\begin{table}[tb]
    \centering
    \resizebox{\columnwidth}{!}{
    \begin{tabular}{l|c|r|c|c}
         \mt variant & Params & Train Time & \textbf{en} (Src) & \bf All\\
         \hline
         first-pool & \multirow{2}{*}{277M} & 32e / 25h & 72.4 & \bf 55.6 \\
         last-pool &  & 18e / 14h & \bf 72.8 & 48.8 \\
         \hline
         generative & \multirow{2}{*}{391M} &  3e / 3h & 2.5 & 2.5 \\
         decode-cls & & 21e / 19h & 72.7 & 52.8 \\
         \hline
         \xlmroberta & 278M & 22e / 12h & 73.1 & 50.4 \\
         \hline
    \end{tabular}
    }
    \caption{Comparing \mt variants. The first two variants use only the encoder; the latter two use both the encoder and the decoder. We show \mrp results (\%) on English development data (Src), and development \mrp averaged over all 23 languages (All). We also report the number of trainable parameters (in millions), and training time in epochs (e) and hours (h).}
    \vspace*{-4mm}
    \label{tab:mt_modes_extra}
\end{table}

\begin{table*}[t]
    \centering
    \resizebox{\textwidth{}}{!}{
    \begin{tabular}{lc|l||l|l}
    \multicolumn{2}{l|}{\bf Language} & \bf Model & \bf Publication & \bf Pretraining Corpora \\
    \hline
    English & (\bf en) & \texttt{bert-base-uncased} & \cite{devlin2019} & Wikipedia + Books \\
    Danish & (\bf da) & \texttt{DJSammy/bert-base-danish-uncased\_BotXO,ai} & - & Wikipedia + Web + Subtitles \\
    German & (\bf de) & \texttt{deepset/gbert-base} & \cite{gbert} & Wikipedia + OSCAR + OPUS \\
    Dutch & (\bf nl) & \texttt{pdelobelle/robbert-v2-dutch-base} & \cite{delobelle2020robbert} & Wikipedia + Books + News \\
    Swedish & (\bf sv) & \texttt{KB/bert-base-swedish-cased} & - & Wikipedia + Books + News \\
    Spanish & (\bf es) & \texttt{BSC-TeMU/roberta-base-bne} & \cite{spanishroberta} & Web \\
    French & (\bf fr) & \texttt{camembert-base} & \cite{camembert} & OSCAR \\
    Italian & (\bf it) & \texttt{dbmdz/bert-base-italian-uncased} & - & Wikipedia + OPUS \\
    Portuguese & (\bf pt) & \texttt{neuralmind/bert-base-portuguese-cased} & \cite{bertimbau} & Web \\
    Czech & (\bf cs) & \texttt{UWB-AIR/Czert-B-base-cased} & \cite{czert} & Wikipedia + Web + News\\
    Romanian & (\bf ro) & \texttt{dumitrescustefan/bert-base-romanian-uncased-v1} & - & Wikipedia + OSCAR + OPUS \\
    Polish & (\bf pl) & \texttt{dkleczek/bert-base-polish-uncased-v1} & - & Wikipedia \\
    Estonian & (\bf et) & \texttt{tartuNLP/EstBERT} & - & Web \\
    Finish & (\bf fi) & \texttt{TurkuNLP/bert-base-finnish-uncased-v1} & \cite{finbert} & Web + News \\
    Hungarian & (\bf hu) & \texttt{SZTAKI-HLT/hubert-base-cc} & \cite{hubert} & Wikipedia + OSCAR \\
    Greek & (\bf el) & \texttt{nlpaueb/bert-base-greek-uncased-v1} & \cite{greekbert} & Wikipedia + OSCAR \\
    \end{tabular}
    }
    \vspace*{-2mm}
    \caption{Monolingual (native) \bert models used. We also report the training corpora used to pretrain each model.}
    \vspace*{-3mm}
    \label{tab:nativebert}
\end{table*}

Table~\ref{tab:mt_modes_extra} reports results on development data. As expected, the \emph{generative} version of \mt performs terribly (\mrp 2.5), as the model tries to learn an unnecessary label ordering; in fact the model cannot learn and stops training after five epochs due to early stopping. By contrast, the \emph{decode-cls} variant, which feeds the decoder only with the \cls token and uses its output embedding, has comparable performance with the encoder-only variants (\emph{first-pool}, \emph{last-pool}). 
It uses, however, approximately 40\% more parameters, because of the additional cross-attention layers in the decoder blocks.

Both encoder-only variants of \mt are comparable with \xlmroberta (English \mrp approx.\ 73; the All \mrp scores are also comparable or better).
These results show that the encoder of \mt can be used alone (without the decoder) for text classification, similarly to \transformer{-based} encoder-only models \cite{devlin2019, liu2019}, despite its text-to-text generative pretraining, unlike the generative fine-tuning proposed by the creators of \mt \cite{xue2021}.

\section{Monolingual \bert Models}
\label{sec:appendix_c}

Table~\ref{tab:nativebert} lists all native \bert models used in the experiments of Section~\ref{sec:mono} in the one-to-one set-up. All models are hosted by Hugging Face (\url{https://huggingface.co/models}). All models follow the \textsc{base} configuration with 12 layers of stacked \transformer{s}, each with $D_h=768$ hidden units and 12 attention heads. We use case sensitive models, when available. We cannot guarantee the quality of the different models, as they come from different sources (organizations or individuals), although we tried to select the best possible options, i.e., those trained on more data for a longer period, in case there were many alternatives. We found 16 monolingual models; 
we found no monolingual models for Bulgarian, Slovak, Croatian, Slovene, Lithuanian, Latvian, Maltese.

Most monolingual \bert models use a vocabulary of approx.\ 30k sub-words and have approx.\ 110M parameters in total (24M for embeddings and 86M for \transformer blocks), while \xlmroberta has a much larger vocabulary of 250k sub-words to support 100 languages and 278M parameters (192M for embeddings and 86M for \transformer blocks). Similarly, \mt uses a vocabulary of equal size, thus its encoder has 86M parameters, while its decoder has 120M parameters; as in the work of \citet{vaswani2017}, the decoder \transformer blocks of \mt have more parameters than the encoder blocs, as they use additional cross-attention layers. Based on the aforementioned details, the encoder's capacity is almost identical across the examined models.


\section{Dataset Card for \multieurlex}
\label{sec:data_card}

\subsection{Dataset Description}

\noindent{\textbf{Documents:}}
\multieurlex comprises 65k \eu laws (published 1958--2016) in 23 official \eu languages (Table~\ref{tab:stats}). Each \eu law has been annotated with \eurovoc concepts (labels) by EU's Publications Office. Each \eurovoc label \textsc{id} is associated with a \emph{label descriptor}, e.g., $\langle$60, `agri-foodstuffs'$\rangle$,  $\langle$6006, `plant product'$\rangle$, $\langle$1115, `fruit'$\rangle$. The descriptors are also available in the 23 languages.\vspace{2mm}

\noindent{\textbf{Languages:}} The \eu has 24 official languages. When new members join the \eu, the set of official languages usually expands, unless the new languages are already included. \multieurlex covers 23 languages from seven language families (Germanic, Romance, Slavic, Uralic, Baltic, Semitic, Hellenic). \eu laws are published in all official languages, except Irish, for resource-related reasons.\footnote{\url{https://europa.eu/european-union/about-eu/eu-languages_en}} This wide coverage makes \multieurlex a valuable testbed for cross-lingual transfer. All languages use the Latin script, except for Bulgarian (Cyrillic script) and Greek. Several other languages are also spoken in \eu countries. The \eu is home to over 60 additional indigenous regional or minority languages, e.g., Basque, Catalan, Frisian, Saami, and Yiddish, among others, spoken by approx.\ 40 million people, but these additional languages are not considered official (in terms of \eu), and \eu laws are not translated to them.\vspace{2mm}

\noindent\textbf{Annotation:} All the documents of the dataset have been annotated by the Publications Office of \eu (\url{https://publications.europa.eu/en}) with multiple concepts from \eurovoc (\url{http://eurovoc.europa.eu/}). \eurovoc has eight levels of concepts. Each document is assigned one or more concepts (labels). If a document is assigned a concept, the ancestors and descendants of that concept are typically not assigned to the same document. The documents were originally annotated with concepts from levels 3 to 8.
We augmented the annotation with three alternative sets of labels per document, replacing each assigned concept by its ancestor from level 1, 2, or 3, respectively. 
Thus, we provide four sets of gold labels per document, one for each of the first three levels of the hierarchy, plus the original sparse label assignment.\footnote{Levels 4 to 8 cannot be used independently, as many documents have gold concepts from the third level; thus many documents will be mislabeled, if we discard level 3.}\vspace{2mm}

\noindent{\textbf{Data Split and Concept Drift:}} 
\multieurlex is \emph{chronologically} split in training (55k), development (5k), test (5k) subsets, using the English documents. The test subset contains the same 5k documents in all 23 languages (Table~\ref{tab:stats}).\footnote{The development subset also contains the same 5k documents in 23 languages, except Croatian. Croatia is the most recent \eu member (2013); older laws are gradually translated.}
For the official languages of the seven oldest member countries, the same 55k training documents are available; for the other languages, only a subset of the 55k training documents is available (Table~\ref{tab:stats}).
Compared to \eurlex \cite{chalkidis2019}, \multieurlex is not only larger (8k more documents) and multilingual; it is also more challenging, as the chronological split leads to temporal real-world \emph{concept drift} across the training, development, test subsets, i.e., differences in label distribution and phrasing, representing a realistic \emph{temporal generalization} problem \cite{lazaridou2021}. \citet{sogaard2021} showed this setup is more realistic, as it does not over-estimate real performance, contrary to random splits \cite{gorman-bedrick-2019-need}.\vspace{2mm}

\noindent{\textbf{Supported Tasks:}} 
\multieurlex can be used for legal topic classification, a multi-label classification task where legal documents need to be assigned concepts reflecting their topics. \multieurlex supports labels from three different granularities (\eurovoc levels). More importantly, apart from monolingual (\emph{one-to-one}) experiments, it can be used to study cross-lingual transfer scenarios, including \emph{one-to-many} (systems trained in one language and used in other languages with no training data), and \emph{many-to-one} or \emph{many-to-many} (systems jointly trained in multiple languages and used in one or more other languages).\vspace{2mm}

\noindent\textbf{Data Fields:} The following data fields are provided for all documents of \multieurlex:

\begin{itemize}[leftmargin=12pt]
\vspace*{-2mm}
    \setlength\itemsep{0.0mm}
    \item `celex\_id`: (\textbf{str})  The official ID of the document. The \textsc{celex} number is the unique identifier for all publications in both \textsc{eur-lex} and \textsc{cellar}, the \eu Publications Office’s common repository of metadata and content.
    \item `publication\_date`: (\textbf{str})  The publication date of the document.
    \item `text`: (\textbf{dict[str]})  A dictionary with (key, value) pairs, where the key is the 2-letter \textsc{iso} code of each language and the value is the content of each document in this language.
  \item `eurovoc\_concepts`: (\textbf{dict[List[str]]}) A dictionary with (key, value) pairs, where the key is the label set (level 1--3) and the value is a list of the relevant \eurovoc concepts (labels).
\end{itemize}

\subsection{Initial Data Collection and Normalization}

The original data are available at the \textsc{eur-lex} portal (\url{https://eur-lex.europa.eu}) in unprocessed formats (\textsc{html}, \textsc{xml}, \textsc{rdf}). The documents were downloaded from the \textsc{eurlex} portal in \textsc{html}. The relevant \eurovoc concepts were downloaded from the \textsc{sparql} endpoint of the Publications Office of \eu (\url{http://publications.europa.eu/webapi/rdf/sparql}). 
We stripped \textsc{html} mark-up to provide the documents in plain text format.
We inferred the labels for \eurovoc levels 1--3, by backtracking the \eurovoc hierarchy branches, from the originally assigned labels to their ancestors in levels 1--3, respectively.

\subsection{Personal and Sensitive Information}

The dataset contains publicly available \eu laws that do not include personal or sensitive information, with the exception of trivial information presented by consent, e.g., the names of the current presidents of the European Parliament and European Council, and other administration bodies.

\subsection{Licensing Information}

We provide \multieurlex with the same licensing as the original \eu data (\textsc{cc-by-4.0}):

\medskip
\noindent\emph{The Commission’s document reuse policy is based on Decision 2011/833/EU. Unless otherwise specified, you can re-use the legal documents published in \textsc{eur-lex} for commercial or non-commercial purposes.} \vspace{2mm}

\noindent \emph{The copyright for the editorial content of this website, the summaries of \eu legislation and the consolidated texts, which is owned by the \eu, is licensed under the Creative Commons Attribution 4.0 International licence. This means that you can re-use the content provided you acknowledge the source and indicate any changes you have made.}\vspace{2mm}

{\small
\noindent\underline{Source:} \url{https://eur-lex.europa.eu/content/legal-notice/legal-notice.html}

\noindent\underline{See also:} \url{https://eur-lex.europa.eu/content/help/faq/reuse-contents-eurlex.html}
}

\section{More Detailed Results}
\label{sec:appendix_a}
For completeness, in Table~\ref{tab:full_results} we present detailed results across all 23 languages for \xlmroberta fine-tuned end-to-end or using the alternative adaptation strategies in the \emph{one-to-many} setting for English. We observe that (a) native \bert models have the best results in 12 out of 15 languages; (b) \xlmroberta trained in a monolingual (one-to-one) setting has competitive results; and (c) fine-tuning with adapter modules leads to the best overall results in cross-lingual transfer and in the many-to-many setting.

Tables~\ref{tab:full_all}--\ref{tab:adapters_all} show the results when fine-tuning end-to-end or using adapters, considering each one of the 23 languages as a source language in a \emph{one-to-many} setting.

Table~\ref{tab:results_gran_all} shows \xlmroberta results for all \eurovoc levels across all 23 languages. 

Table~\ref{tab:full_results_mt5} reports detailed results across all 23 languages for the alternative adaptation strategies using the \emph{first-pool} \mt variant.\vspace{2mm}

Similarly to Table~\ref{fig:data_splits}, Table~\ref{tab:data_splits_2} shows the effects of temporal concept drift in the performance of \xlmroberta, for Level 3 with 567 labels.

\begin{table}[h]
    \centering
    \resizebox{0.9\columnwidth}{!}{
    \begin{tabular}{l|c|c|c}
    \hline
    Data Split & Training & Development & Test \\
    \hline
    Random &  \bf 93.0 & \bf 80.9 & \bf 80.3 \\
    Chronological & 92.8 &  73.1 &  67.4 \\
    \hline
    \end{tabular}
    }
    \caption{Results of \multieurlex for level 3 (567 labels) with \xlmroberta using a \emph{random} or \emph{chronological} split. Here the model is fine-tuned and tested on English data only (\emph{one-to-one}).}
    \label{tab:data_splits_2}
    \vspace{-5mm}
\end{table}

\begin{sidewaystable*}
    \centering
    \resizebox{\textwidth}{!}{
    \begin{tabular}{lc|cccc|ccccc|cccccc|ccc|cc|c|c|c}
        \hline
           & \multicolumn{5}{c}{\textsc{Germanic}} & \multicolumn{5}{c}{\textsc{Romance}} & \multicolumn{6}{c}{\textsc{Slavic}} & \multicolumn{3}{c}{\textsc{Uralic}} & \multicolumn{2}{c}{\textsc{Baltic}} & \multicolumn{2}{c}{} \\
         & \bf en   &	 \bf da   &	 \bf de   &	 \bf nl   &	 \bf sv   &	 \bf ro   &	 \bf es   &	 \bf fr   &	 \bf it   &	 \bf pt   &	 \bf bg   &	 \bf cs   &	 \bf hr   &	 \bf pl   &	 \bf sk   &	 \bf sl   &	 \bf hu   &	 \bf fi   &	 \bf et   &	 \bf lt   &	 \bf lv   &	 \bf el   &	 \bf mt   &	 All \\
         \hline
        \multicolumn{24}{l}{\textbf{One-to-one} (Fine-tune \xlmroberta or monolingually pretrained \bert{s} in one language, test in the \emph{same} language.)} \\
        \hline
        \nativebert & 67.7 & 65.5 & 68.4 & 66.7 & 68.5 & 68.5 & 67.6 & 67.4 & 67.9 & 67.4 & - & 66.7 & - & 67.2 & - & - & 67.7 & 67.8 & 66.0 & - & - & 67.8 & - & 67.4 \\
        \xlmroberta & 67.4 & 66.7 & 67.5 & 67.3 & 66.5 & 66.4 & 67.8 & 67.2 & 67.4 & 67.0 & 66.1 & 66.7 
        & 61.7 & 65.0 & 64.8 & 66.7 & 65.5 & 66.5 & 65.7 & 66.2 & 66.7 & 65.8 & 62.9 & 66.6 \\
        Diff. 	& -0.3 & +1.2 & -0.9 & +0.6 & -2.0 & -2.1 & +0.2 & -0.2 & -0.5 & -0.4 & - & 0.0 & - & -2.2 & - & - & -2.2 & -1.3 & -0.3 & - & - & -2.0 & - & -0.7 \\
        \hline
        \multicolumn{24}{l}{\textbf{One-to-many} (Fine-tune \xlmroberta \emph{only} in English, test in all languages, with alternative adaptation strategies.)} \\
        \hline
        End-to-end fine-tuning & 67.4 & 56.5 & 52.4 & 49.0 & 55.7 & 55.2 & 54.0 & 55.0 & 52.0 & 50.5 & 51.2 & 49.6 & 49.6 & 46.9 & 49.3 & 49.9 & 48.8 & 46.4 & 45.2 & 49.7 & 46.4 & 33.3 & 20.4 & 49.3 \\
        First three blocks frozen & 66.3 &	 59.1 &	 56.8 &	 55.3 &	 57.5 &	 57.9 &	 58.1 &	 57.7 &	 56.2 &	 54.9 &	 56.1 &	 54.3 &	 52.8 &	 53.7 &	 53.0 &	 51.4 &	 51.0 &	 52.1 &	 49.7 &	 51.3 &	 50.1 &	 42.4 &	 20.3 &	 53.0 \\
        First six blocks frozen  & 66.3 &	 59.1 &	 57.4 &	 55.7 &	 57.9 &	 57.2 &	 56.9 &	 57.9 &	 53.9 &	 55.4 &	 55.8 &	 52.6 &	 49.2 &	 51.9 &	 50.8 &	 49.3 &	 47.3 &	 48.7 &	 45.0 &	 48.5 &	 49.9 &	 39.6 &	 22.0 &	 51.7 \\
        First nine blocks frozen  & 65.8 &	 59.4 &	 57.9 &	 56.9 &	 58.6 &	 58.2 &	 58.7 &	 59.4 &	 55.7 &	 57.5 &	 56.7 &	 54.2 &	 50.7 &	 53.4 &	 54.4 &	 48.7 &	 48.8 &	 50.4 &	 46.2 &	 51.6 &	 50.5 &	 44.5 &	 21.4 &	 53.0 \\
        All 12 blocks frozen & 27.2 & 21.4 & 24.6 & 24.6 & 23.0 & 21.6 & 23.4 & 21.9 & 20.1 & 25.1 & 23.1 & 24.3 & 19.9 & 22.8 & 26.0 & 19.8 & 22.8 & 21.9 & 20.2 & 22.9 & 21.4 & 19.0 & 14.2 & 22.2 \\
        Adapters layers  & 67.3 & 61.5 & 59.3 & 57.8 & 59.5 & 60.3 & 61.0 & 60.4 & 58.8 & 58.5 & 59.2 & 56.8 & 56.9 & 57.5 & 57.0 & 53.5 & 55.3 & 55.6 & 53.1 & 55.2 & 52.4 & 46.1 & 27.4 & 56.1 \\
        \bitfit (bias terms only) & 63.9 & 59.3 & 57.0 & 54.0 & 58.2 & 57.8 & 57.4 & 56.9 & 56.4 & 55.5 & 55.6 & 54.8 & 55.1 & 54.0 & 52.8 & 57.9 & 51.2 & 54.8 & 52.3 & 52.5 & 51.8 & 42.1 & 22.7 & 53.7 \\
        \hline
        \multicolumn{18}{l}{\textbf{Many-to-many} (Jointly fine-tune \xlmroberta in \emph{all} languages, test in all languages, with alternative adaptation strategies.)} \\
        \hline
        End-to-end fine-tuning & 66.4 & 66.2 & 66.2 & 66.1 & 66.1 & 66.3 & 66.3 & 66.2 & 66.3 & 65.9 & 65.7 & 65.7 & 65.8 & 65.6 & 65.7 & 65.8 & 65.2 & 65.8 & 65.6 & 65.7 & 65.8 & 65.1 & 62.3 & 65.7 \\
        Adapters layers & \bf 67.3 & \bf 67.1 & \bf 66.3 & \bf 67.1 & \bf 67.0 & \bf 67.4 & \bf 67.2 & \bf 67.1 & \bf 67.4 & \bf 67.0 & \bf 66.6 & \bf 67.0 & \bf 67.0 & \bf 66.2 & \bf 66.2 & \bf 66.8 & \bf 65.5 & \bf 66.6 & \bf 65.7 & \bf 65.8 & \bf 66.7 & \bf 65.7 & \bf 61.6 &  \bf 66.4 \\
    \end{tabular}
    }
    \caption{Test results for \xlmroberta in cross-lingual classification at level 3 (567 labels). We show \mrp (\%) for each one of the 23 languages, and \mrp averaged over all 23 languages. 
    }
    \label{tab:full_results}
    \vspace{-4mm}
\end{sidewaystable*}

\begin{sidewaystable*}
    \centering
    \resizebox{\textwidth}{!}{
    \begin{tabular}{lc|cccc|ccccc|cccccc|ccc|cc|c|c|c}
        \hline
           & \multicolumn{5}{c}{\textsc{Germanic}} & \multicolumn{5}{c}{\textsc{Romance}} & \multicolumn{6}{c}{\textsc{Slavic}} & \multicolumn{3}{c}{\textsc{Uralic}} & \multicolumn{2}{c}{\textsc{Baltic}} & \multicolumn{2}{c}{} \\
         & \bf en   &	 \bf da   &	 \bf de   &	 \bf nl   &	 \bf sv   &	 \bf ro   &	 \bf es   &	 \bf fr   &	 \bf it   &	 \bf pt   &	 \bf bg   &	 \bf cs   &	 \bf hr   &	 \bf pl   &	 \bf sk   &	 \bf sl   &	 \bf hu   &	 \bf fi   &	 \bf et   &	 \bf lt   &	 \bf lv   &	 \bf el   &	 \bf mt   &	 All \\
         \hline
         \bf en & 67.4 & 56.5 & 52.4 & 49.0 & 55.7 & 55.2 & 54.0 & 55.0 & 52.0 & 50.5 & 51.2 & 49.6 & 49.6 & 46.9 & 49.3 & 49.9 & 48.8 & 46.4 & 45.2 & 49.7 & 46.4 & 33.3 & 20.4 & 49.3 \\
         \bf da & 55.6 & 66.7 & 54.5 & 53.6 & 58.3 & 50.8 & 50.2 & 48.6 & 47.7 & 47.1 & 49.7 & 46.9 & 47.1 & 44.1 & 46.6 & 47.3 & 46.5 & 42.4 & 43.1 & 47.0 & 43.1 & 33.6 & 17.7 & 47.3 \\
         \bf de & 55.3 & 56.9 & 67.5 & 52.9 & 55.9 & 50.3 & 49.4 & 48.6 & 46.2 & 45.3 & 48.1 & 49.0 & 47.9 & 46.5 & 50.2 & 46.5 & 49.6 & 44.5 & 44.0 & 43.5 & 40.5 & 34.9 & 17.5 & 47.4 \\
         \bf nl & 55.9 & 56.5 & 55.2 & 67.3 & 53.1 & 50.1 & 50.8 & 50.0 & 48.1 & 47.5 & 49.3 & 49.8 & 49.5 & 48.7 & 50.6 & 46.8 & 45.6 & 45.3 & 42.4 & 47.1 & 44.2 & 35.9 & 18.2 & 48.2 \\
         \bf sv & 58.3 & 61.5 & 55.0 & 52.5 & 66.5 & 52.2 & 51.1 & 50.7 & 49.2 & 49.9 & 51.7 & 50.0 & 50.2 & 48.4 & 50.4 & 49.3 & 47.5 & 47.9 & 48.3 & 49.8 & 45.0 & 37.0 & 18.8 & 49.6 \\
         \bf ro & 57.7 & 53.2 & 51.9 & 50.5 & 52.5 & 66.4 & 59.8 & 57.4 & 56.2 & 54.1 & 53.4 & 50.7 & 50.9 & 51.2 & 50.6 & 52.6 & 51.0 & 46.2 & 49.2 & 50.2 & 45.9 & 40.4 & 18.3 & 50.9 \\
         \bf es & 58.6 & 53.8 & 49.8 & 48.3 & 52.6 & 59.3 & 67.8 & 58.9 & 59.3 & 61.5 & 52.5 & 47.8 & 49.3 & 48.6 & 47.4 & 46.8 & 47.2 & 41.6 & 41.4 & 44.2 & 43.4 & 36.4 & 19.3 & 49.4 \\
         \bf fr & 60.6 & 55.0 & 52.1 & 54.1 & 53.9 & 58.4 & 60.3 & 67.2 & 57.8 & 58.5 & 53.9 & 50.9 & 52.8 & 51.0 & 50.8 & 50.5 & 49.6 & 46.0 & 45.4 & 47.8 & 44.6 & 38.1 & 22.1 & 51.4 \\
         \bf it & 56.7 & 52.3 & 46.1 & 45.4 & 50.2 & 56.5 & 60.7 & 56.9 & 67.4 & 56.6 & 51.3 & 46.2 & 47.9 & 43.7 & 47.5 & 47.6 & 44.1 & 40.6 & 42.0 & 42.8 & 40.1 & 33.9 & 22.8 & 47.8 \\
         \bf pt & 59.0 & 54.5 & 50.5 & 48.6 & 52.2 & 57.9 & 63.0 & 59.2 & 58.8 & 67.0 & 51.9 & 47.5 & 50.2 & 48.5 & 49.0 & 51.1 & 49.5 & 45.2 & 42.3 & 46.3 & 43.2 & 35.9 & 20.7 & 50.1 \\
         \bf bg & 55.8 & 51.5 & 51.1 & 48.3 & 50.5 & 53.5 & 52.0 & 51.2 & 47.9 & 48.7 & 66.1 & 50.4 & 53.3 & 50.6 & 50.2 & 51.0 & 48.3 & 46.4 & 40.7 & 50.1 & 45.6 & 39.6 & 14.7 & 48.6 \\
         \bf cs & 53.1 & 52.4 & 50.5 & 48.0 & 51.2 & 49.1 & 52.0 & 48.9 & 48.9 & 49.1 & 51.7 & 66.7 & 55.9 & 51.7 & 61.7 & 55.8 & 50.5 & 46.3 & 44.2 & 49.4 & 44.7 & 36.0 & 16.1 & 49.3 \\
         \bf hr & 50.4 & 50.6 & 49.1 & 48.1 & 50.8 & 52.8 & 50.5 & 49.9 & 50.6 & 47.9 & 51.4 & 51.8 & 61.7 & 51.4 & 53.3 & 54.6 & 47.0 & 49.1 & 46.8 & 47.9 & 46.5 & 37.0 & 19.6 & 48.6 \\
         \bf pl & 53.6 & 50.8 & 48.5 & 47.7 & 50.2 & 51.3 & 52.1 & 49.1 & 45.4 & 49.1 & 52.2 & 51.7 & 51.6 & 65.0 & 51.9 & 49.8 & 45.6 & 42.1 & 39.5 & 44.4 & 41.3 & 35.3 & 13.1 & 47.0 \\
         \bf sk & 53.8 & 53.2 & 52.4 & 49.7 & 51.2 & 50.9 & 52.1 & 50.8 & 49.1 & 48.4 & 53.6 & 60.4 & 54.4 & 53.5 & 64.8 & 54.1 & 49.4 & 47.2 & 44.9 & 49.2 & 45.0 & 38.1 & 16.0 & 49.7 \\
         \bf sl & 54.2 & 53.4 & 49.5 & 50.0 & 52.4 & 52.2 & 53.1 & 51.5 & 51.8 & 49.9 & 53.3 & 54.6 & 58.1 & 51.0 & 53.3 & 66.7 & 48.7 & 47.7 & 45.5 & 47.8 & 45.8 & 38.7 & 15.0 & 49.7 \\
         \bf hu & 51.9 & 49.7 & 47.5 & 42.9 & 46.9 & 48.9 & 48.6 & 47.1 & 45.2 & 43.8 & 48.8 & 47.0 & 46.8 & 46.5 & 47.0 & 46.3 & 65.5 & 44.3 & 43.9 & 44.2 & 41.3 & 33.4 & 14.2 & 45.3 \\
         \bf fi & 47.6 & 48.5 & 43.7 & 41.9 & 48.2 & 44.7 & 43.3 & 42.0 & 43.8 & 38.7 & 42.7 & 42.6 & 45.2 & 38.5 & 43.3 & 45.6 & 43.9 & 66.5 & 45.2 & 43.7 & 38.7 & 30.3 & 11.2 & 42.6 \\
         \bf et & 51.2 & 49.5 & 44.8 & 44.6 & 48.7 & 45.8 & 48.2 & 43.9 & 44.6 & 42.6 & 43.4 & 44.6 & 44.9 & 42.6 & 44.6 & 45.4 & 46.1 & 46.2 & 65.7 & 45.2 & 43.4 & 32.2 & 14.7 & 44.5 \\
         \bf lt & 54.3 & 48.4 & 46.0 & 41.9 & 49.3 & 49.8 & 48.9 & 47.4 & 46.5 & 43.4 & 48.2 & 46.0 & 48.4 & 46.7 & 47.2 & 49.3 & 45.2 & 41.2 & 43.9 & 66.2 & 49.0 & 32.2 & 16.9 & 45.9 \\
         \bf lv & 51.8 & 49.4 & 45.9 & 48.1 & 49.3 & 51.3 & 50.6 & 49.8 & 47.8 & 48.4 & 51.1 & 48.9 & 48.9 & 49.2 & 49.6 & 49.5 & 49.0 & 44.3 & 48.3 & 53.0 & 66.7 & 38.1 & 18.8 & 48.1 \\
         \bf el & 41.2 & 37.8 & 34.3 & 33.3 & 38.0 & 38.2 & 42.3 & 38.5 & 39.6 & 41.3 & 39.3 & 34.5 & 38.8 & 32.1 & 36.8 & 35.6 & 36.8 & 31.3 & 28.9 & 34.7 & 30.3 & 65.8 &  9.6 & 36.5 \\
         \bf mt & 27.5 & 28.4 & 23.2 & 25.3 & 25.5 & 25.6 & 25.3 & 24.8 & 22.9 & 23.9 & 24.1 & 22.2 & 22.0 & 21.9 & 23.0 & 22.1 & 23.6 & 21.0 & 22.9 & 23.9 & 23.2 & 19.6 & 62.9 & 25.4 \\

    \end{tabular}
    }
    \caption{Test results (mRP, \%) for Level 3 (567 labels) with \xlmroberta, when fine-tuning end-to-end in one language (source, rows) and testing in all languages (columns).}
    \label{tab:full_all}
    \vspace{-4mm}
\end{sidewaystable*}

\begin{sidewaystable*}
    \centering
    \resizebox{\textwidth}{!}{
    \begin{tabular}{lc|cccc|ccccc|cccccc|ccc|cc|c|c|c}
        \hline
           & \multicolumn{5}{c}{\textsc{Germanic}} & \multicolumn{5}{c}{\textsc{Romance}} & \multicolumn{6}{c}{\textsc{Slavic}} & \multicolumn{3}{c}{\textsc{Uralic}} & \multicolumn{2}{c}{\textsc{Baltic}} & \multicolumn{2}{c}{} \\
         & \bf en   &	 \bf da   &	 \bf de   &	 \bf nl   &	 \bf sv   &	 \bf ro   &	 \bf es   &	 \bf fr   &	 \bf it   &	 \bf pt   &	 \bf bg   &	 \bf cs   &	 \bf hr   &	 \bf pl   &	 \bf sk   &	 \bf sl   &	 \bf hu   &	 \bf fi   &	 \bf et   &	 \bf lt   &	 \bf lv   &	 \bf el   &	 \bf mt   &	 All \\
         \hline
         \bf en & 66.8 & 61.5 & 59.3 & 57.8 & 59.5 & 60.3 & 61.0 & 60.4 & 58.8 & 58.5 & 59.2 & 56.8 & 56.9 & 57.5 & 57.0 & 53.5 & 55.3 & 55.6 & 53.1 & 55.2 & 52.4 & 46.1 & 27.4 & 56.1 \\
         \bf da & 62.0 & 66.6 & 60.1 & 59.6 & 62.4 & 58.4 & 58.7 & 57.1 & 54.5 & 56.4 & 58.1 & 57.2 & 56.9 & 55.4 & 57.9 & 55.0 & 56.6 & 55.9 & 52.5 & 55.8 & 52.7 & 46.5 & 21.2 & 55.5 \\
         \bf de & 60.7 & 61.0 & 67.1 & 58.5 & 59.8 & 57.7 & 57.8 & 56.9 & 56.1 & 55.5 & 56.7 & 56.1 & 55.8 & 55.9 & 56.2 & 54.4 & 55.8 & 53.5 & 51.7 & 51.3 & 51.3 & 44.8 & 24.5 & 54.8 \\
         \bf nl & 61.6 & 62.0 & 59.9 & 67.4 & 59.2 & 58.7 & 58.5 & 58.1 & 55.0 & 56.1 & 57.6 & 58.2 & 57.4 & 56.1 & 58.2 & 54.0 & 55.7 & 53.7 & 50.8 & 54.3 & 52.0 & 45.4 & 23.4 & 55.4 \\
         \bf sv & 61.8 & 63.8 & 60.4 & 58.2 & 66.9 & 57.1 & 57.8 & 56.5 & 55.4 & 55.8 & 57.3 & 56.8 & 55.3 & 55.1 & 57.1 & 55.3 & 53.7 & 55.0 & 52.5 & 54.4 & 51.5 & 44.9 & 22.7 & 55.0 \\
         \bf ro & 62.0 & 59.8 & 59.3 & 58.1 & 58.0 & 67.0 & 60.9 & 59.8 & 58.9 & 58.8 & 59.5 & 56.5 & 56.8 & 57.4 & 56.8 & 58.2 & 56.1 & 55.0 & 55.9 & 56.8 & 55.3 & 48.4 & 29.2 & 56.7 \\
         \bf es & 62.8 & 59.2 & 56.5 & 57.3 & 57.1 & 62.7 & 67.7 & 61.7 & 61.8 & 63.0 & 58.3 & 57.0 & 55.9 & 56.6 & 57.5 & 53.1 & 54.1 & 51.8 & 51.1 & 50.6 & 52.3 & 45.3 & 23.1 & 55.5 \\
         \bf fr & 63.7 & 60.8 & 60.6 & 58.6 & 58.0 & 61.6 & 63.0 & 67.4 & 60.0 & 60.3 & 59.7 & 57.9 & 59.2 & 58.2 & 58.9 & 55.4 & 56.9 & 55.2 & 51.7 & 53.6 & 53.1 & 47.3 & 31.4 & 57.1 \\
         \bf it & 61.1 & 58.2 & 53.4 & 57.0 & 55.2 & 60.1 & 61.5 & 60.8 & 67.4 & 60.6 & 54.9 & 52.9 & 54.2 & 53.6 & 51.5 & 54.2 & 51.4 & 53.6 & 51.0 & 52.4 & 51.9 & 47.7 & 26.5 & 54.4 \\
         \bf pt & 63.9 & 59.1 & 57.1 & 57.5 & 56.8 & 61.4 & 64.9 & 60.6 & 61.2 & 67.6 & 59.2 & 56.0 & 57.9 & 56.1 & 56.6 & 55.4 & 54.6 & 52.6 & 49.6 & 53.4 & 53.0 & 47.1 & 24.4 & 55.9 \\
         \bf bg & 61.2 & 58.3 & 58.5 & 56.3 & 57.7 & 59.0 & 57.4 & 57.5 & 54.7 & 56.1 & 66.9 & 58.3 & 58.7 & 58.4 & 58.5 & 53.7 & 56.6 & 53.3 & 51.2 & 53.8 & 50.9 & 47.8 & 23.3 & 55.1 \\
         \bf cs & 59.8 & 60.1 & 58.0 & 56.8 & 56.9 & 57.5 & 57.8 & 58.5 & 56.6 & 56.1 & 59.0 & 67.2 & 60.5 & 59.4 & 64.6 & 59.0 & 57.2 & 54.8 & 54.8 & 56.1 & 55.4 & 46.9 & 25.7 & 56.5 \\
         \bf hr & 56.0 & 52.4 & 52.5 & 51.7 & 54.8 & 54.3 & 53.8 & 54.5 & 53.0 & 51.6 & 55.4 & 54.7 & 62.4 & 52.7 & 55.6 & 55.3 & 49.9 & 50.8 & 49.4 & 52.3 & 50.1 & 41.6 & 24.6 & 51.7 \\
         \bf pl & 59.2 & 58.8 & 57.4 & 55.8 & 56.7 & 59.2 & 57.3 & 57.4 & 53.7 & 56.4 & 58.8 & 58.5 & 57.6 & 66.5 & 59.7 & 56.0 & 55.1 & 54.9 & 48.6 & 54.3 & 53.2 & 45.6 & 20.6 & 54.8 \\
         \bf sk & 60.0 & 59.5 & 58.0 & 57.7 & 58.3 & 59.3 & 58.6 & 57.8 & 55.5 & 54.5 & 59.7 & 62.5 & 60.4 & 58.6 & 66.6 & 59.6 & 56.3 & 55.3 & 55.7 & 55.4 & 55.9 & 46.4 & 23.4 & 56.3 \\
         \bf sl & 60.8 & 59.3 & 57.9 & 57.6 & 59.2 & 59.1 & 59.1 & 58.0 & 56.1 & 56.7 & 60.5 & 59.5 & 62.5 & 58.2 & 60.0 & 66.0 & 56.1 & 56.5 & 54.2 & 56.0 & 55.6 & 48.3 & 24.8 & 56.6 \\
         \bf hu & 61.0 & 59.6 & 59.7 & 57.6 & 57.5 & 57.6 & 59.3 & 56.9 & 55.1 & 55.2 & 59.2 & 57.9 & 57.5 & 56.9 & 56.9 & 56.0 & 66.2 & 55.9 & 54.4 & 54.9 & 54.3 & 47.9 & 23.7 & 55.7 \\
         \bf fi & 58.6 & 57.3 & 54.7 & 53.8 & 56.5 & 56.8 & 56.8 & 53.9 & 54.5 & 54.2 & 54.4 & 54.9 & 53.3 & 53.0 & 53.2 & 53.2 & 55.2 & 66.8 & 54.4 & 51.0 & 48.8 & 43.1 & 21.9 & 53.1 \\
         \bf et & 58.8 & 57.0 & 54.8 & 54.4 & 57.4 & 57.0 & 56.0 & 54.1 & 54.4 & 52.8 & 55.8 & 55.0 & 55.6 & 54.4 & 55.3 & 53.3 & 57.1 & 54.9 & 66.2 & 54.6 & 53.7 & 44.0 & 21.0 & 53.8 \\
         \bf lt & 59.2 & 57.6 & 54.1 & 53.7 & 56.6 & 56.2 & 55.1 & 56.7 & 54.1 & 51.4 & 56.8 & 55.4 & 53.3 & 54.4 & 56.0 & 53.4 & 55.0 & 55.2 & 52.2 & 67.0 & 56.5 & 43.8 & 27.0 & 53.9 \\
         \bf lv & 58.6 & 57.1 & 56.4 & 53.7 & 57.7 & 56.8 & 56.4 & 54.7 & 54.3 & 53.5 & 57.2 & 55.4 & 55.0 & 55.3 & 55.9 & 54.9 & 55.5 & 55.2 & 55.2 & 58.1 & 66.5 & 45.4 & 22.3 & 54.4 \\
         \bf el & 46.8 & 48.9 & 47.8 & 49.5 & 49.3 & 48.8 & 52.0 & 50.6 & 48.9 & 49.4 & 51.0 & 50.1 & 48.6 & 48.4 & 49.9 & 47.7 & 48.2 & 46.2 & 41.5 & 44.2 & 45.1 & 65.5 & 20.6 & 47.8 \\
         \bf mt & 43.7 & 44.3 & 40.7 & 39.7 & 40.7 & 42.8 & 44.5 & 42.6 & 45.3 & 42.0 & 38.7 & 39.1 & 40.2 & 38.4 & 40.0 & 39.1 & 38.6 & 35.2 & 35.9 & 40.5 & 39.7 & 32.8 & 63.9 & 41.2 \\
    \end{tabular}
    }
    \caption{Test results (mRP, \%) for Level 3 (567 labels) with \xlmroberta fine-tuned with adapter modules in one language (source, rows) and testing in all languages (columns).}
    \label{tab:adapters_all}
    \vspace{-4mm}
\end{sidewaystable*}

\begin{sidewaystable*}
    \centering
    \resizebox{\textwidth}{!}{
    \begin{tabular}{lc|cccc|ccccc|cccccc|ccc|cc|c|c|c}
        \hline
           & \multicolumn{5}{c}{\textsc{Germanic}} & \multicolumn{5}{c}{\textsc{Romance}} & \multicolumn{6}{c}{\textsc{Slavic}} & \multicolumn{3}{c}{\textsc{Uralic}} & \multicolumn{2}{c}{\textsc{Baltic}} & \multicolumn{2}{c}{} \\
         & \bf en   &	 \bf da   &	 \bf de   &	 \bf nl   &	 \bf sv   &	 \bf ro   &	 \bf es   &	 \bf fr   &	 \bf it   &	 \bf pt   &	 \bf bg   &	 \bf cs   &	 \bf hr   &	 \bf pl   &	 \bf sk   &	 \bf sl   &	 \bf hu   &	 \bf fi   &	 \bf et   &	 \bf lt   &	 \bf lv   &	 \bf el   &	 \bf mt   &	 All \\
         \hline
        \multicolumn{24}{l}{\textbf{One-to-many} (Fine-tune \mt \emph{only} in English, test in all languages, with alternative adaptation strategies.)} \\
         \hline
    End-to-end fine-tuning       & 67.4 & 59.5 & 58.0 & 57.1 & 58.9 & 58.5 & 58.3 & 59.6 & 54.9 & 54.8 & 56.2 & 52.3 & 40.7 & 52.2 & 50.9 & 53.2 & 52.6 & 51.1 & 51.3 & 53.3 & 51.4 & 43.5 & 39.5 & 53.7 \\
    First 3 blocks frozen      & 67.4 & \bf 62.6 & 60.0 & \bf 60.5 & \bf 61.0 & 61.4 & 61.3 & 60.9 & \bf 60.0 & 58.2 & \bf 58.8 & 56.1 & 45.4 & 54.9 & 56.4 & 56.9 & \bf 54.5 & 54.2 & 55.4 & 56.2 & 56.4 & 46.7 & 44.0 & 56.9 \\
    First 6 blocks frozen       & 66.3 & 61.8 & 59.3 & 61.1 & 61.7 & 61.0 & 61.5 & 61.7 & 58.9 & 59.5 & 60.9 & 57.9 & 48.5 & 57.8 & 57.9 & 59.4 & 56.2 & 58.7 & 59.2 & 58.9 & 57.1 & 50.9 & 46.3 & 58.4 \\
    First 9 blocks frozen       & \bf 68.0 & 61.9 & \bf 60.8 & 59.1 & \bf 61.0 & \bf 63.1 & \bf 63.2 & \bf 63.7 & 59.1 & \bf 61.8 & 58.7 & \bf 58.4 & \bf 47.1 & \bf 56.8 & \bf 58.2 & \bf 59.4 & 54.0 & \bf 55.6 & \bf 57.2 & \bf 58.1 & \bf 57.5 & \bf 48.5 & \bf 48.7 & \bf 58.3 \\
    Adapters layers   & 66.3 & 53.0 & 48.8 & 47.1 & 49.3 & 48.3 & 53.8 & 52.9 & 49.8 & 48.4 & 45.5 & 41.7 & 28.2 & 43.1 & 35.9 & 38.0 & 41.2 & 42.5 & 35.8 & 38.4 & 40.4 & 37.2 & 27.5 & 44.0 \\
    \end{tabular}
    }
    \caption{Test results for \mt (first-pool) in cross-lingual classification at level 3 (567 labels). We show \mrp (\%) for each one of the 23 languages, and \mrp averaged over all 23 languages.
    }
    \label{tab:full_results_mt5}
    \vspace{-4mm}
\end{sidewaystable*}

\begin{sidewaystable*}
    \centering
    \resizebox{\textwidth}{!}{
    \begin{tabular}{llc|cccc|ccccc|cccccc|ccc|cc|c|c|c}
\eurovoc & & \multicolumn{5}{c}{\textsc{Germanic}} & \multicolumn{5}{c}{\textsc{Romance}} & \multicolumn{6}{c}{\textsc{Slavic}} & \multicolumn{3}{c}{\textsc{Uralic}} & \multicolumn{2}{c}{\textsc{Baltic}} & \multicolumn{2}{c}{} \\
         Label Set & (\#Labels) & \bf en   &	 \bf da   &	 \bf de   &	 \bf nl   &	 \bf sv   &	 \bf ro   &	 \bf es   &	 \bf fr   &	 \bf it   &	 \bf pt   &	 \bf bg   &	 \bf cs   &	 \bf hr   &	 \bf pl   &	 \bf sk   &	 \bf sl   &	 \bf hu   &	 \bf fi   &	 \bf et   &	 \bf lt   &	 \bf lv   &	 \bf el   &	 \bf mt   &	 All \\
         \hline
         \multicolumn{24}{c}{\textbf{One-to-many} (Fine-tune \xlmroberta end-to-end \emph{only} in English, test in all languages.)} \\
         \hline
         Level 1 & (21) & 83.2 & 78.0 & 78.5 & 75.9 & 75.8 & 77.4 & 78.7 & 78.6 & 77.2 & 77.5 & 78.3 & 76.6 & 76.7 & 76.8 & 76.0 & 76.2 & 75.5 & 74.5 & 75.0 & 76.2 & 75.3 & 71.4 & 52.9 & 75.7 \\
         Level 2 & (127) & 73.6 & 64.1 & 61.2 & 58.8 & 61.0 & 64.5 & 65.6 & 63.0 & 61.8 & 60.3 & 61.1 & 59.0 & 59.1 & 58.8 & 58.1 & 60.5 & 58.5 & 55.4 & 55.4 & 57.8 & 55.1 & 47.2 & 31.3 & 58.7 \\
         Level 3 & (567) & 67.4 & 56.5 & 52.4 & 49.0 & 55.7 & 55.2 & 54.0 & 55.0 & 52.0 & 50.5 & 51.2 & 49.6 & 49.6 & 46.9 & 49.3 & 49.9 & 48.8 & 46.4 & 45.2 & 49.7 & 46.4 & 33.3 & 20.4 & 49.3 \\
         All & (7,390) & 43.0 & 28.5 & 26.9 & 25.4 & 30.6 & 31.5 & 29.2 & 30.5 & 30.2 & 30.1 & 28.4 & 21.6 & 25.1 & 24.5 & 22.2 & 24.5 & 20.5 & 20.9 & 18.8 & 19.2 & 17.3 & 14.9 & 4.7 & 24.7 \\
         \hline
         \multicolumn{24}{c}{\textbf{One-to-many} (Fine-tune \xlmroberta with adapter modules \emph{only} in English, test in all languages.)} \\
         \hline
         Level 1 & (21) & 83.1 & 80.3 & 79.1 & 78.9 & 77.9 & 80.3 & 78.0 & 78.6 & 79.1 & 78.4 & 79.7 & 77.7 & 77.5 & 77.5 & 77.6 & 77.9 & 76.0 & 76.2 & 76.5 & 76.7 & 78.5 & 75.3 & 54.8 & 77.2 \\
         Level 2 & (127) & 73.2 & 65.9 & 66.6 & 62.3 & 62.0 & 66.9 & 65.2 & 67.0 & 62.0 & 64.1 & 64.3 & 61.4 & 65.0 & 60.1 & 62.4 & 64.5 & 61.6 & 59.4 & 56.8 & 58.0 & 61.2 & 49.4 & 31.2 & 61.3 \\
        Level 3 & (567) & 66.8 & 61.5 & 59.3 & 57.8 & 59.5 & 60.3 & 61.0 & 60.4 & 58.8 & 58.5 & 59.2 & 56.8 & 56.9 & 57.5 & 57.0 & 53.5 & 55.3 & 55.6 & 53.1 & 55.2 & 52.4 & 46.1 & 27.4 & 56.1 \\
         All & (7,390)  & 42.8 & 37.0 & 32.7 & 34.5 & 36.2 & 36.7 & 35.7 & 33.5 & 33.4 & 36.7 & 35.7 & 35.0 & 33.0 & 32.7 & 34.5 & 32.6 & 31.6 & 30.1 & 31.6 & 32.2 & 30.2 & 27.3 & 13.4 & 33.0 \\
         \hline
    \end{tabular}
    }
    \caption{Test results of \xlmroberta fine-tuned \emph{end-to-end} or with adapters across all \eurovoc levels (label sets).}
    \label{tab:results_gran_all}
    \vspace{-4mm}
\end{sidewaystable*}

\begin{sidewaystable*}
    \centering
    \resizebox{\textwidth}{!}{
    \begin{tabular}{lccccc|ccccc|cccccc|ccc|cc|c|c|c}
     & \multicolumn{5}{c}{\textsc{Germanic}} & \multicolumn{5}{c}{\textsc{Romance}} & \multicolumn{6}{c}{\textsc{Slavic}} & \multicolumn{3}{c}{\textsc{Uralic}} & \multicolumn{2}{c}{\textsc{Baltic}} & \multicolumn{2}{c}{} \\
    \bf Family (Src)	& \bf en   &	 \bf da   &	 \bf de   &	 \bf nl   &	 \bf sv   &	 \bf ro   &	 \bf es   &	 \bf fr   &	 \bf it   &	 \bf pt   &	 \bf bg   &	 \bf cs   &	 \bf hr   &	 \bf pl   &	 \bf sk   &	 \bf sl   &	 \bf hu   &	 \bf fi   &	 \bf et   &	 \bf lt   &	 \bf lv   &	 \bf el   &	 \bf mt   &	 All \\
    \hline
    \multicolumn{24}{c}{\textbf{Many-to-many} (Fine-tune \xlmroberta end-to-end in \emph{all languages of the same family}, test in all languages.)} \\
    \hline
    \textsc{Germanic} & 67.9 & 67.8 & 67.6 & 66.1 & 67.4 & 61.1 & 60.2 & 60.3 & 58.2 & 58.9 & 60.6 & 58.0 & 61.2 & 58.1 & 58.4 & 59.4 & 57.0 & 56.3 & 52.9 & 57.9 & 56.5 & 45.9 & 26.4 & 58.4 \\
    \textsc{Romance}  & 65.2 & 58.5 & 54.8 & 56.7 & 57.7 & 67.1 & 67.3 & 67.4 & 67.6 & 66.0 & 60.9 & 56.0 & 57.3 & 55.3 & 55.1 & 56.5 & 54.0 & 48.3 & 50.3 & 51.0 & 49.5 & 40.0 & 30.9 & 56.2 \\
    \textsc{Slavic}  & 63.8 & 60.7 & 62.0 & 61.6 & 60.5 & 62.9 & 61.5 & 61.6 & 59.7 & 59.3 & 67.4 & 66.6 & 67.1 & 65.6 & 65.0 & 65.6 & 57.7 & 59.2 & 58.6 & 60.7 & 59.0 & 49.0 & 23.2 & 59.9 \\
    \textsc{Uralic}	 & 60.4 & 56.1 & 55.2 & 50.4 & 55.7 & 56.3 & 55.0 & 53.4 & 50.4 & 50.9 & 52.1 & 52.9 & 56.1 & 49.0 & 53.1 & 53.1 & 65.5 & 66.0 & 49.4 & 53.0 & 50.2 & 39.8 & 16.1 & 52.2 \\
    \textsc{Baltic}  & 59.8 & 53.3 & 47.6 & 51.9 & 53.5 & 53.4 & 53.2 & 52.2 & 51.2 & 51.0 & 55.2 & 53.5 & 54.7 & 52.7 & 54.8 & 57.4 & 51.0 & 48.2 & 53.0 & 67.3 & 66.7 & 36.1 & 18.9 & 52.0 \\
    \hline
    \multicolumn{24}{c}{\textbf{Many-to-many} (Fine-tune \xlmroberta with adapter modules in \emph{all languages of the same family}, test in all languages.)} \\
    \hline
    \textsc{Germanic} & 67.6 & 67.5 & 67.4 & 66.8 & 67.0 & 65.3 & 65.2 & 62.7 & 60.3 & 62.1 & 64.2 & 64.0 & 62.3 & 63.5 & 64.0 & 62.1 & 62.1 & 62.4 & 59.7 & 59.6 & 62.2 & 52.4 & 30.4 & 61.8 \\
    \textsc{Romance}  & 66.6 & 65.6 & 62.9 & 63.3 & 61.3 & 66.1 & 67.9 & 67.8 & 67.7 & 67.5 & 64.8 & 61.8 & 60.6 & 60.8 & 61.9 & 61.3 & 62.0 & 59.5 & 59.6 & 59.1 & 61.0 & 54.8 & 33.5 & 61.6 \\
    \textsc{Slavic}  & 63.6 & 63.0 & 62.7 & 61.5 & 64.0 & 62.9 & 61.8 & 62.4 & 57.9 & 59.1 & 66.0 & 66.4 & 64.8 & 66.2 & 66.3 & 66.5 & 61.6 & 61.4 & 56.0 & 61.6 & 60.6 & 53.1 & 33.6 & 61.0 \\
    \textsc{Uralic}   & 61.5 & 60.3 & 59.3 & 59.4 & 60.2 & 57.9 & 59.3 & 58.1 & 56.1 & 57.7 & 60.3 & 58.5 & 59.3 & 57.9 & 56.7 & 58.1 & 66.5 & 66.7 & 58.2 & 57.9 & 56.4 & 49.4 & 28.6 & 57.6 \\
    \textsc{Baltic}   & 62.4 & 60.0 & 57.2 & 59.2 & 59.8 & 60.1 & 60.9 & 58.1 & 58.2 & 58.5 & 60.8 & 56.4 & 62.9 & 60.0 & 56.7 & 58.9 & 57.5 & 58.9 & 56.3 & 66.0 & 65.9 & 51.2 & 30.7 & 58.1 \\
         \hline
    \end{tabular}
    }
    \caption{Test results (\mrp, \%) for Level 3 (567 labels) with \xlmroberta, when fine-tuning \emph{end-to-end} or with adapters in \emph{all languages of the same family (Src)}, test in all languages. We show \mrp (\%) for each one of the 23 languages, and \mrp averaged over all 23 languages.}
    \label{tab:families_2}
    \vspace{-4mm}
\end{sidewaystable*}

\end{document}